%% file: main_aistats2022.tex
\documentclass[twoside]{article}

\usepackage[accepted]{headers/aistats2022}

\usepackage[round]{natbib}

\usepackage{xpatch}
\usepackage{array}
\usepackage{hyperref} %
\usepackage{algorithm}
\usepackage{algorithmic}
\usepackage{listing}
\usepackage{graphicx}
\usepackage{subcaption}
\usepackage{longtable}
\usepackage{supertabular}
\usepackage{booktabs}
\usepackage{float}
\usepackage[export]{adjustbox}
\usepackage{breakcites}
\usepackage{color}
\usepackage{marginnote}
\usepackage{fancyvrb}
\usepackage{listings}
\usepackage{wrapfig}
\usepackage{sidecap}
\usepackage[noabbrev,capitalise]{cleveref}

\usepackage{microtype,xfrac}

\input{./math_definition}
\input{./personal_commands}

\usepackage[scaled=0.9]{sourcecodepro}
\begin{document}

\runningtitle{Policy Learning and Evaluation with RQMC}

\runningauthor{Arnold, L'Ecuyer, Chen, Chen, Sha}

\twocolumn[
    \aistatstitle{
        Policy Learning and Evaluation \\
        with Randomized Quasi-Monte Carlo
    }
    \aistatsauthor{S\'ebastien M. R. Arnold \And Pierre L'Ecuyer \And  Liyu Chen}
    \aistatsaddress{\texttt{seb.arnold@usc.edu}\\U. of Southern California \And \texttt{lecuyer@iro.umontreal.ca}\\U. de Montr\'eal \And \texttt{liyuc@usc.edu}\\U. of Southern California}
    \aistatsauthor{Yi-fan Chen \And Fei Sha}
    \aistatsaddress{\texttt{yifanchen@google.com}\\Google Research \And \texttt{fsha@google.com}\\Google Research}
]

\begin{abstract}
    Reinforcement learning constantly deals with hard integrals, for example when computing expectations in policy evaluation and policy iteration.
    These integrals are rarely analytically solvable and typically estimated with the Monte Carlo method, which induces high variance in policy values and gradients.
    In this work, we propose to replace Monte Carlo samples with low-discrepancy point sets.
    We combine policy gradient methods with Randomized Quasi-Monte Carlo, yielding variance-reduced formulations of policy gradient and actor-critic algorithms.
    These formulations are effective for policy evaluation and policy improvement, as they outperform state-of-the-art algorithms on standardized continuous control benchmarks.
    Our empirical analyses validate the intuition that replacing Monte Carlo with Quasi-Monte Carlo yields significantly more accurate gradient estimates.
\end{abstract}

\input{intro}

\input{background}

\input{method}

\input{experiments}

\input{related}

\input{conclusion}

\subsubsection*{Acknowledgements}
We thank Florian Wenzel for stimulating discussion in the early stages of the project.
We appreciate the feedback from the reviewers.
This work is partially supported by NSF Awards IIS-1513966/ 1632803/1833137, CCF-1139148, DARPA Award\#: FA8750-18-2-0117,  FA8750-19-1-0504, DARPA-D3M - Award UCB-00009528, Google Research Awards, gifts from Facebook and Netflix, and ARO\# W911NF-12-1-0241 and W911NF-15-1-0484.

\bibliographystyle{apalike}
\bibliography{main_aistats2022}

\clearpage
\appendix
\thispagestyle{empty}
\onecolumn
\input{supp/supp_aistats}

\end{document}

%% file: math_definition.tex
\usepackage{amssymb}
\usepackage{amsmath,amsfonts}
\usepackage{amsopn,bm,mathtools}

\usepackage{nicefrac}       %

\newcommand{\field}[1]{\mathbb{#1}}
\newcommand{\R}{\field{R}} %

\newcommand{\ProbOpr}[1]{\mathbb{#1}}

\newcommand{\expect}[2]{%
\ifthenelse{\equal{#2}{}}{\ProbOpr{E}_{#1}}
{\ifthenelse{\equal{#1}{}}{\ProbOpr{E}\left[#2\right]}{\ProbOpr{E}_{#1}\left[#2\right]}}} %
\newcommand{\var}[2]{%
\ifthenelse{\equal{#2}{}}{\ProbOpr{VAR}_{#1}}
{\ifthenelse{\equal{#1}{}}{\ProbOpr{VAR}\left[#2\right]}{\ProbOpr{VAR}_{#1}\left[#2\right]}}} %

%% file: personal_commands.tex
\usepackage{amsthm}
\usepackage{amsmath}
\usepackage{amssymb}
\usepackage{float}
\usepackage{xspace}
\usepackage[dvipsnames]{xcolor}

\usepackage[colorinlistoftodos,linecolor=blue,backgroundcolor=white,bordercolor=blue,textsize=tiny,textwidth=22mm]
{todonotes}
\let\olditemize=\itemize
\let\endolditemize=\enditemize
\renewenvironment{itemize}{
    \olditemize
    \setlength{\itemsep}{0pt}
    \setlength{\parskip}{0pt}
}{\endolditemize}

\let\oldenumerate=\enumerate
\let\endoldenumerate=\endenumerate
\renewenvironment{enumerate}{
    \oldenumerate
    \setlength{\itemsep}{0pt}
    \setlength{\parskip}{0pt}
}{\endoldenumerate}

\newcommand{\paren}[1]{\left( #1 \right)}
\newcommand{\brackets}[1]{\left[ #1 \right]}

\newcommand{\norm}[1]{\left\lVert#1\right\rVert}

\newcommand{\bigO}[1]{\mathcal{O}\paren{#1}}

\DeclareMathOperator*{\E}{\mathop{\mathbb{E}}}
\newcommand{\Exp}[2][]{\E_{#1}\brackets{#2}}

\newcommand{\supp}{Supp. Material\xspace}
\newcommand{\mujoco}{MuJoCo\xspace}
\newcommand{\swimmer}{\texttt{Swimmer-v3}\xspace}
\newcommand{\cheetah}{\texttt{HalfCheetah-v3}\xspace}
\newcommand{\ant}{\texttt{Ant-v3}\xspace}
\newcommand{\hopper}{\texttt{Hopper-v3}\xspace}
\newcommand{\walker}{\texttt{Walker2d-v3}\xspace}

\newcommand{\eg}{\emph{e.g.}\xspace}
\newcommand{\ie}{\emph{i.e.}\xspace}
\newcommand{\params}{{\theta}}
\newcommand{\policy}{{\pi_{\params}}}
\newcommand{\Q}{{Q^\policy}}
\newcommand{\hQ}{{\hat{Q}^\policy}}
\newcommand{\pvalue}{V^{\policy}\xspace}
\newcommand{\arqmc}{\emph{Array}-RQMC\xspace}
\newcommand{\dimA}{\text{dim}(\mathcal{A})}

%% file: intro.tex
\section{INTRODUCTION}

\begin{figure}[th]
        \makebox[0.97\linewidth]{
            \includegraphics[width=0.52\linewidth]{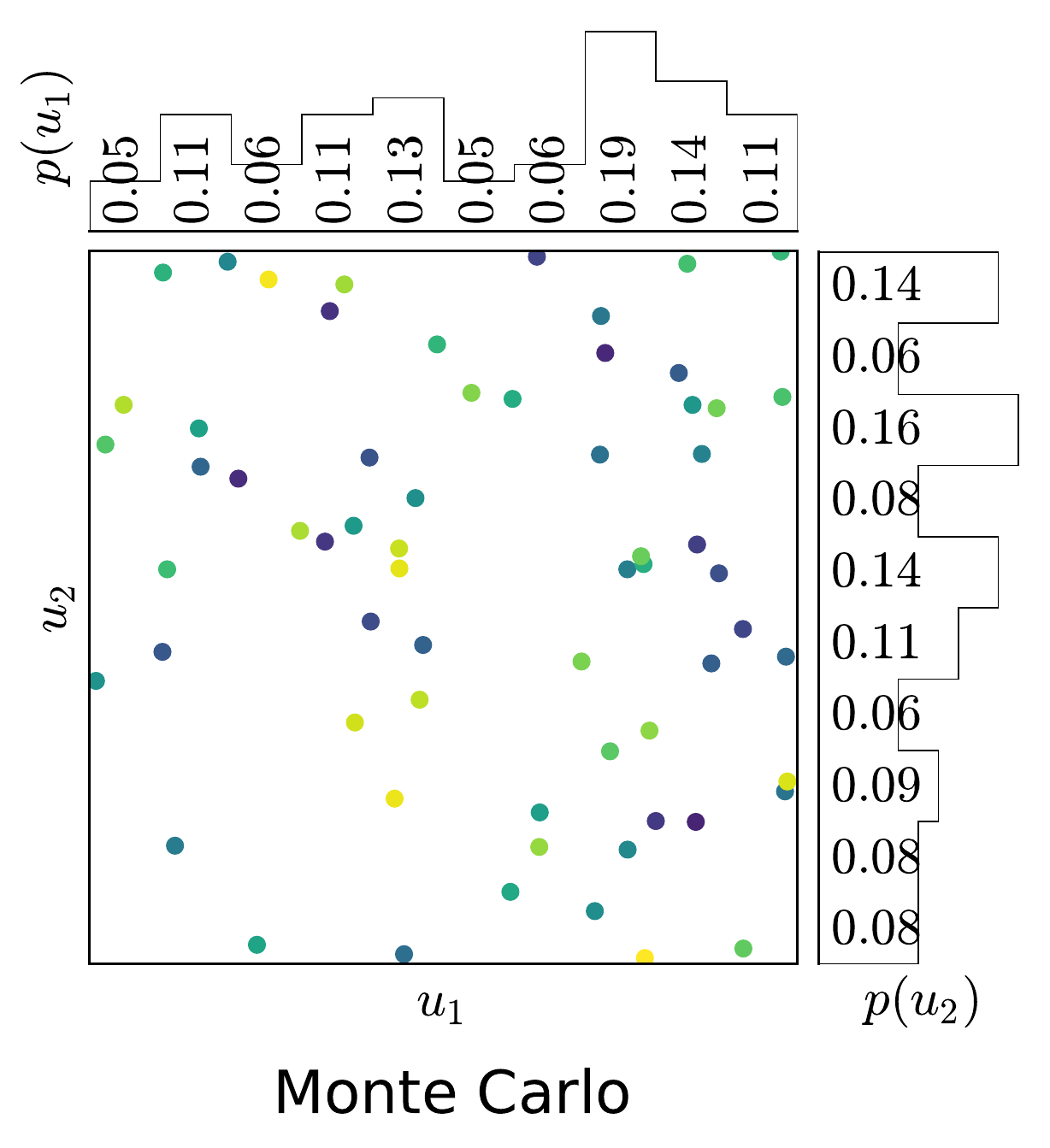}
            \includegraphics[width=0.52\linewidth]{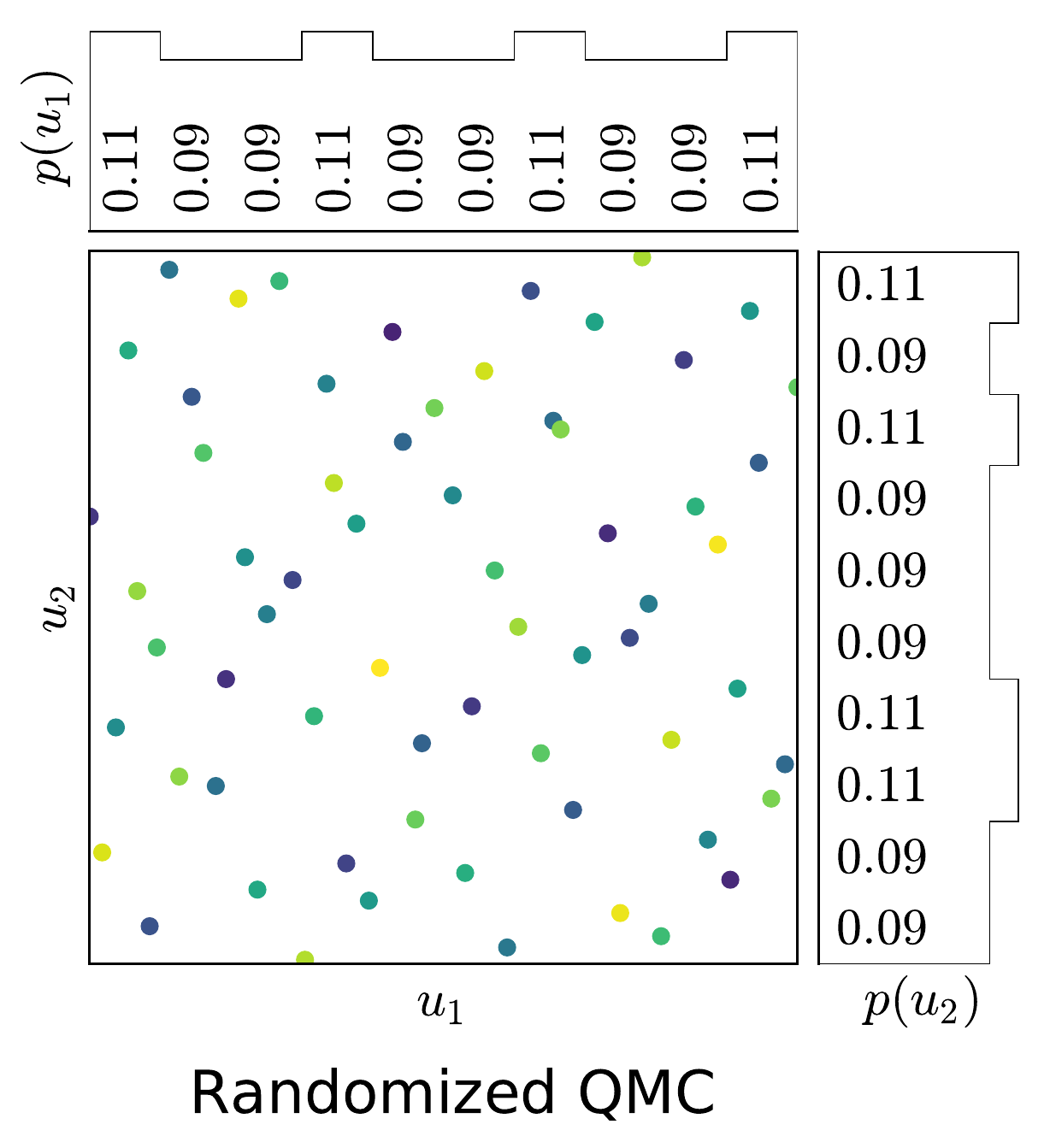}
        }
    \caption{\small 
        \textbf{Comparing MC and RQMC point sets.}
        64 points drawn uniformly over the space $[0, 1)^2$ with Monte-Carlo (MC) and randomized Quasi-Monte Carlo (RQMC). 
        Randomized QMC point sets minimize discrepancy thus covering the space more evenly, as indicated by the histograms of the marginals.
        These point sets improve the estimation of the expectation in \cref{eq:mc}, under some smoothness assumptions.
    }
    \label{fig:pointsets}
    \vspace{-0.2in}
\end{figure}

Policy gradient methods are widely used in reinforcement learning for their broad applicability. They can be applied to optimizing linear or nonlinear policies on non-differentiable and stochastic objectives. In particular, they  handle naturally high-dimensional and continuous action spaces~\citep{benbrahim1997biped,peters2012policy,schulman2015trust,Rajeswaran-RSS-18}

At the core of those methods is to evaluate policies and improve them. Various formulations require to compute some integrals over state and action spaces that are often analyticaly intractable.  In practice, those integrals are approximated and  Monte Carlo (MC) sampling is one of the popular approaches to do so~\citep{Metropolis1949-qy}. This approximation is inherently stochastic; the exact choice and implementation of MC variants directly impacts the quality of the estimated policy value and policy gradients, due to the variance in the approximation. A standard way to battle the variance is to increase the number of MC samples, thus the computational cost as well as the need for the agents to interact with the environments (or their simulators).  The vanilla MC sampling, often implemented in practice as drawing from a random number generator (of a known distribution such as uniform or Gaussian), has a convergence rate of $\bigO{N^{\nicefrac{-1}{2}}}$, where $N$ is the number of samples. 

In this work we explore the benefit of using Randomized Quasi-Monte Carlo (RQMC), a type of Quasi Monte Carlo (QMC) methods as a drop-in replacement of the vanilla sampler~\citep{Niederreiter1978-im}. The key difference of QMC from the standard MC is that QMC uses deterministic point sets to approximate the integration. Under appropriate conditions, including that the integrated function has bounded Hardy-Krause variation, QMC can asymptotically improve the convergence rate from $\bigO{N^{\nicefrac{-1}{2}}}$ to nearly $\bigO{N^{-1}}$~\citep{Papageorgiou2003-dq}.
RQMC improves the efficiency when multiple point sets are needed -- a scenario we will encounter when we need to compute confidence intervals of policy values or compute policy gradients. 

\cref{fig:pointsets} demonstrates the difference between the vanilla MC and RQMC with 2-d point sets. The RQMC point set covers the unit square more evenly than the MC's. Our empirical studies on several tasks demonstrate clearly the utility of this property in improving the estimation of both policy value and gradients.

Other types of methods for reducing variance in estimation have also been explored: control variates such as optimal baselines \citep{greensmith2004variance,10.5555/2100584.2100627,10.5555/2074022.2074088} and value functions \citep{mnih2016asynchronous,schulman2015high}, carefully designed optimization methods~\citep{frostig2015competing,Le_Roux2012-zm, Defazio2014-if,Shalev-Shwartz2013-th}. They are orthogonal to how to obtain samples for integration, and our work also explore the possibility of combining them with RQMC to gain further improvement in estimation quality.

Our main contribution is the introduction of the approach and the empirical studies of its performance and analysis.
Our empirical results establish that RQMC always outperforms MC on both policy evaluation and improvement.
In particular, RQMC improves sample complexity while reducing  policy evaluation error at an order of magnitude.
Further analysis shows that the strong performance of RQMC is due to  reduced gradient variances and better alignment with the true gradients.
RQMC is competitive with other variance reduction techniques, and we show it can be effectively combined with those other techniques to obtain best performance.

The rest of the paper is organized as follows.  We introduce the basic idea behind Quasi-Monte Carlo methods and more specifically, RQMC~(\S\ref{sec:background}). We show how to apply RQMC for policy evaluation and policy improvement~(\S\ref{sec:qmcpg}). In \S\ref{sec:experiments}, we empirically study the performance of MC and RQMC for  policy evaluation and policy improvement on continuous control problems (\eg, Brownian motion, LQR, and \mujoco).

%% file: background.tex
\section{BACKGROUND}\label{sec:background}

Monte Carlo methods (MC) have been used widely in  computational statistics for estimating stochastic integrals. Consider the problem of estimating the expectation of a function $f$ over the $d$-dimensional uniform distribution $U^d$ on the unit cube $[0, 1)^d$.
We obtain an MC estimate by averaging the value of $f$ for $N$ vectors $u^{(i)} \in \R^d$ sampled from $U^d$:
\begin{align}\label{eq:mc}
    \Exp[u \sim U^d]{f(u)} \approx \frac{1}{N} \sum_{i=1}^N f(u^{(i)}).
\end{align}
It is well-known that the mean-squared error of this MC estimator converges at a rate of $\bigO{N^{\nicefrac{-1}{2}}}$~\citep{Robert2004MonteCS}.

\paragraph{Quasi Monte Carlo (QMC)}  The basic idea behind all QMC methods is to improve the standard MC convergence rate by replacing the random samples $u^{(i)}$ with a \emph{deterministic, low discrepancy} point set.
Intuitively, we construct this low-discrepancy point set so as to maximize the distance between each point $u^{(i)}$, thus providing a more even coverage of $[0, 1)^d$, see~ \cref{fig:pointsets} for an illustration.

QMC with such low-discrepancy point sets can reach rates of convergence arbitrarily close to $\bigO{N^{-1}}$ when $f$ satisfies some regularity conditions~\citep{Papageorgiou2003-dq}.
In comparison, common variance-reduction techniques in reinforcement learning (\eg, control variates or importance sampling) only improve the constant factors in front of the $\bigO{N^{\nicefrac{-1}{2}}}$ rate~\citep{Glynn2002-yd, Elvira2021-gv}.

We briefly present a concrete way -- ``digital nets in base 2'' -- to construct point sets as described by~\citet[\S~2.3]{LEcuyer2016-hj}.
Other constructions such as stratification and lattice rules are detailed in~\citep{Niederreiter_1992-gy}.
For $N = 2^k$ points in $d$ dimensions, we begin by choosing $d$  matrices $C_1, \dots, C_d \in [0, 1]^{k \times k}$, \eg, the upper-triangular non-singular ones generating the Sobol sequence~\citep{Sobol1976-dd}, as described by~\citet{joe2008constructing}.
To obtain the $i^\text{th}$ point $u^{(i)}$, we proceed in three steps.
First, we obtain the binary vector representation of $i$:  $b^{(i)} = [b^{(i)}_1, \dots, b^{(i)}_k]$ such that $i  = b^{(i)}_1 + b^{(i)}_2 2 + \dots + b^{(i)}_k 2^{k-1}$ for $i\in [1, N]$.
Second, we multiply this representation with each generating matrix $C_j$:
\begin{equation*}
    \hat{u}_j^{(i)} = C_j b^{(i)} \mod 2, \qquad j = 1, \dots, d,
\end{equation*}
where the modulus is applied to each element of vector $\hat{u}_j^{(i)} \in \mathbb{R}^{k \times 1}$.
Third, we map this base-2 representation back to its decimal form: $u^{(i)}_j = \sum_{l=1}^{k} \hat{u}^{(i)}_{j, l} 2^{-l}$.
The final point $u^{(i)} = [u_1^{(i)}, \dots, u_d^{(i)}]$ is the concatenation of all dimension values $u_1^{(i)}, \dots, u_d^{(i)}$.  Note that the procedure is deterministic as the Sobol sequence (and its generating matrices) is determinstic --- given the generating matrices, the QMC point set is unique.

\begin{figure}[t]
    \begin{center}
        \includegraphics[width=0.32\linewidth]{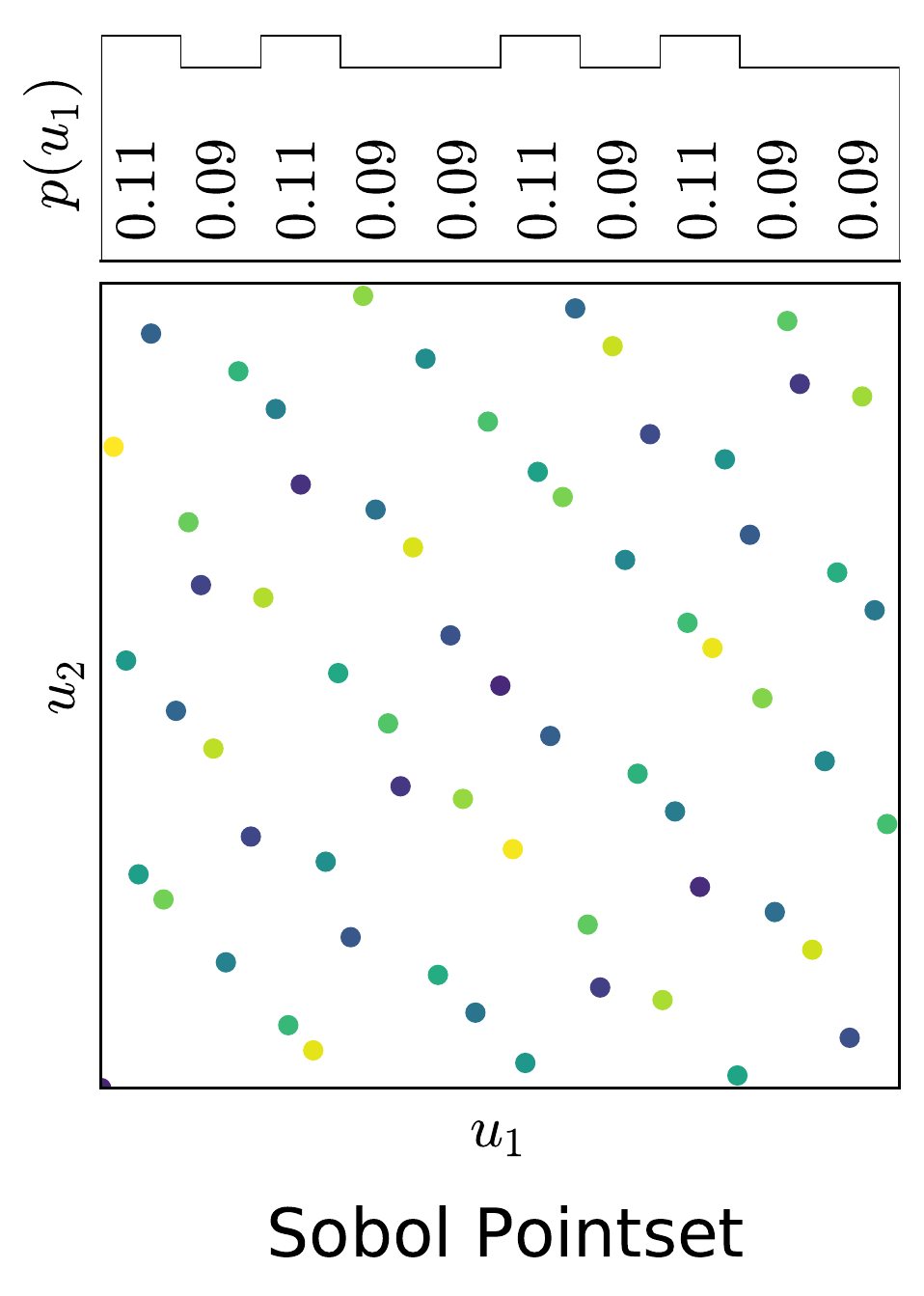}
        \includegraphics[width=0.32\linewidth]{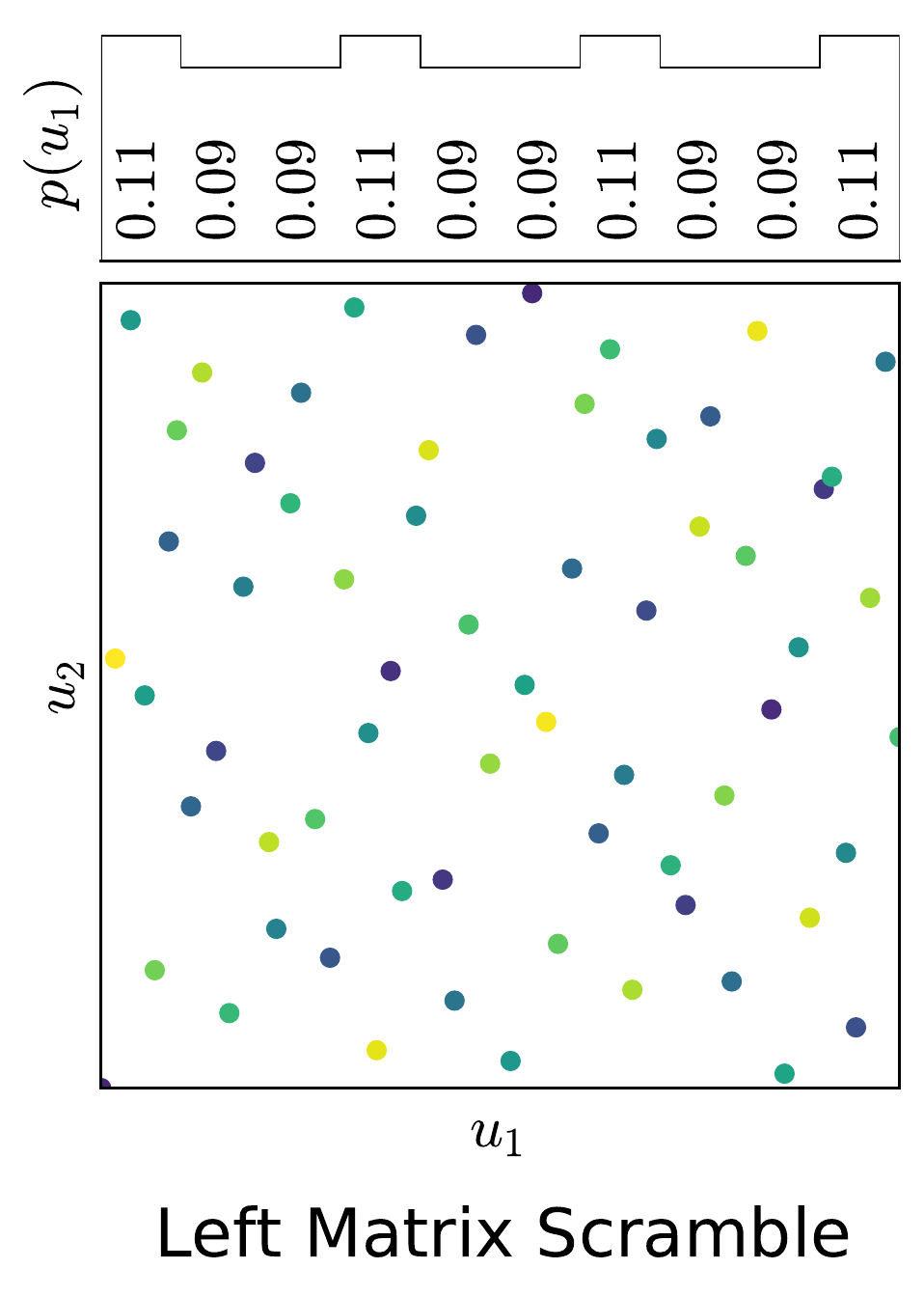}
        \includegraphics[width=0.32\linewidth]{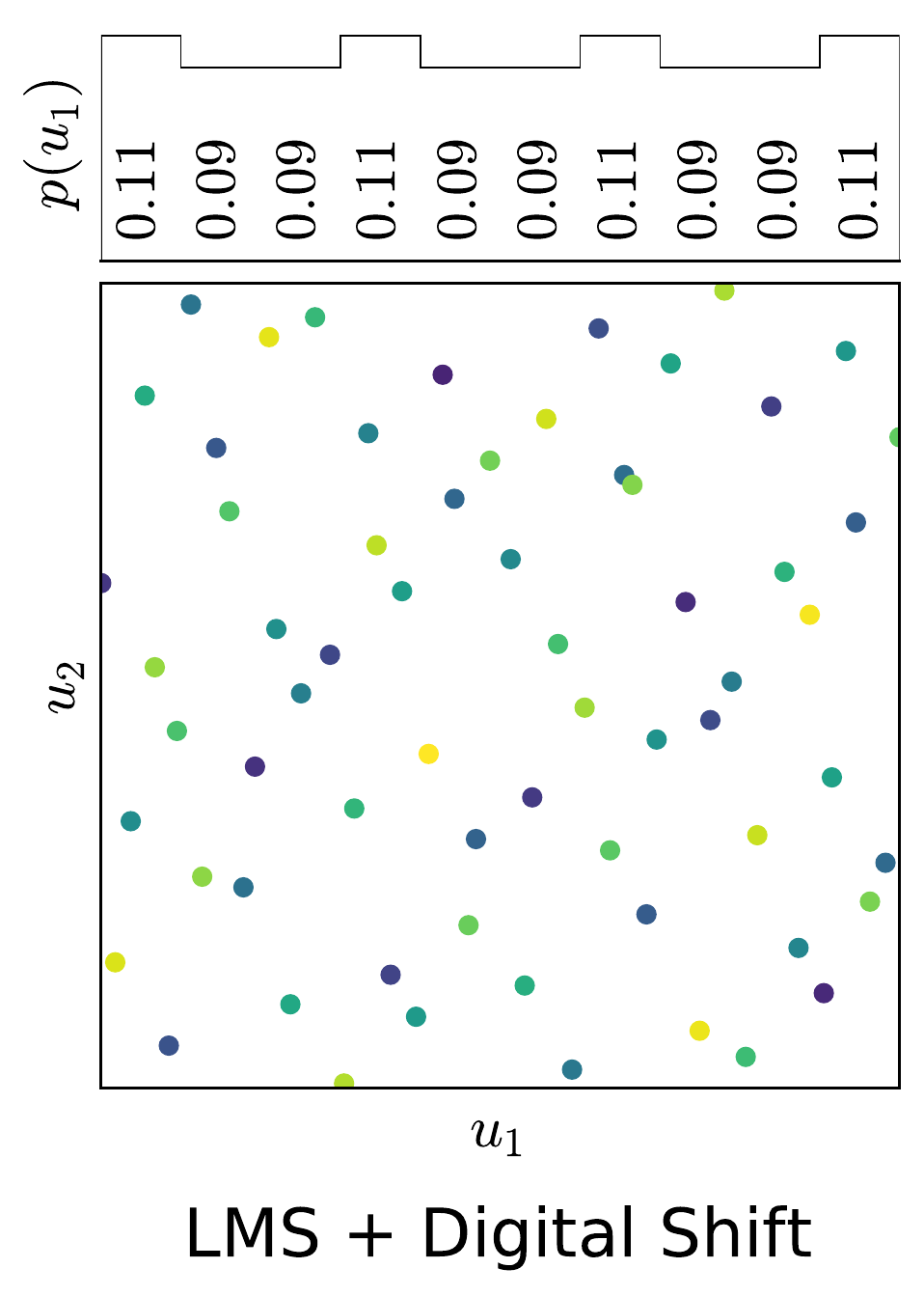}
    \end{center}
    \caption{\small 
        \textbf{Randomized point set Pipeline}
        QMC point sets (Sobol) are generated deterministically to minimize discrepancy, and cover the space more evenly. 
        When several estimates are required, it is common to randomize the point set: first, by applying a left matrix scramble, then a digital shift to debias the point set.
    }
    \label{fig:rqmc-pipeline}
\end{figure}

\paragraph{Randomized QMC (RQMC)}
If we need two or more independent realizations of the expectation $\Exp[u]{f(u)}$ (\eg, for confidence intervals or iterative procedures), we can randomize QMC such that the resulting point set retains low-discrepancy while also yielding unbiased estimates~\citep{Owen1998-ad, LEcuyer2016-hj}. 

How to best randomize QMC point sets depends on the properties of $f$.
For example, if $f$ is sufficiently smooth, the convergence rate of RQMC can be improved to be close to $\bigO{N^{-3/2}}$~\citep{Owen1997-dy, Matousek1998-yx, Hickernell2000-tv}.
Our presentation and experiments focus on randomization via \emph{left matrix scrambling} followed by a \emph{random digital shift}, as first introduced by~\citet{Matousek1998-yx}.
The operations are intuitively just random linear map and shift.

With the left matrix scramble, we scramble the point set by transforming generating matrices $C_1, \dots, C_d$ with a set of random binary matrices $L_1, \dots, L_d$.
To that end, we first sample $d$ lower-triangular, non-singular, random matrices $L_j \in [0, 1]^{k \times k}$, $j = 1, \dots, d$ and multiply the generating matrices $C_j$:
\[
C_j \gets L_j C_j \mod 2, \qquad j = 1, \dots, d,
\]
where the modulus is applied element-wise.
We then generate our point set as for QMC, using the newly scrambled generating matrices $C_1, \dots, C_d$.
While scrambling effectively and efficiently randomizes the point set, repeating this procedure over multiple point sets results in biased estimates (\eg, $u^{(i)}_j = 0$ remains $0$ across all point sets).
Fortunately, this bias is easily addressed by computing the bit-wise exclusive-or of a random binary  vector $v_j \in [0, 1]^k$ to shift  each point of the point set: 
\begin{equation*}
\hat{u}^{(i)}_j \gets \hat{u}^{(i)}_j \oplus v_j, \qquad j = 1, \dots, d,
\end{equation*}
The procedure is illustrated in \cref{fig:rqmc-pipeline}; we refer the reader to \citep{LEcuyer2016-hj} for a more detailed exposition, and to the \supp for interfaces with popular software packages.

\paragraph{Policy Gradient Methods}
Policy gradient methods learn a parameterized policy $\policy$ by computing the gradient of an objective function with respect to the policy parameters $\params$.
This objective function is  the expected value of the policy, denoted by 
\begin{equation}
  \pvalue =  \Exp[s, a]{\Q(s, a)},
\end{equation}
and defined over the states $s\in \mathcal{S}$ and actions $a \in \mathcal{A} = \mathbb{R}^{\dimA}$ of a Markov decision process.
The states $s$ are sampled according to the visitation frequency while sampling the actions $a$ from the policy.
While in principle the policy can be any distribution, our work assumes a multivariate Gaussian, suitable for continuous control:
\begin{align}
    a \sim \policy(s, u) 
      = \mu_\params(s) + \sigma_\params(s) \odot F^{-1}(u),
      \label{eq:reparam-trick}
\end{align}
where $\mu_\params(s)$ and $\sigma_\params^2(s)$ are the (parameterized) mean and diagonal covariance of the Gaussian for the state $s$, $\odot$ denotes element-wise product, $F^{-1}$ is the inverse cumulative density function of the standard normal distribution, and $u \sim U^{\dimA}$ is a uniformly distributed random variable.
Other policy distributions can be obtained via inverse transform sampling or with the probability integral transform.
When clear from context, we omit the dependency of $\policy$ on $u$ and let $\policy(a \mid s)$ denote the density for taking $a$ in state $s$.

The policy gradient theorem~\citep{sutton2000policy} states that the gradient of the policy value can be computed as:
\begin{align}
    \nabla \pvalue =  
     \Exp[s, a]{Q^\policy(s, a) \nabla_\params \log \policy(a \mid s)},
    \label{eq:rl-pg}
\end{align}
where the right-hand side is obtained by applying the score-function estimator~\citep{Miller1967-mv, Rubinstein1969-sx} on the policy's action-value function $\hQ$.

To estimate the gradient,  the agent needs to interact with the environment by collecting trajectories of state and actions $(s_1, a_1, \dots, s_T, a_T)$. 
$\Q$ can be estimated as expected returns:
\begin{equation}
    \Q(s_k, a_k) = \Exp[s_t, a_t]{\sum\nolimits_{t=k}^{T} R(s_{t}, a_{t})},
\end{equation}
where $R(s, a)$ is the immediate reward observed when taking action $a$ in state $s$.

Another approach, used in actor-critic methods, learns the action-value function (\ie, the critic) $\hQ$ with $\policy$, for example, by minimizing the on-policy squared Bellman error: 
\begin{equation}
    \Exp[s, a, s^\prime]{\left(\Exp[a^\prime]{\hQ(s^\prime, a^\prime)} + R(s, a) - \hQ(s, a)\right)^2},
\label{eq:bellman}
\end{equation}
where the next-state action $a^\prime \sim \policy(\cdot \mid s^\prime)$ is sampled from the current policy, and $s, a, s^\prime$ from a (possibly off-policy) replay buffer. The critic $\hQ$ is then used in computing the policy gradient.

%% file: method.tex
\begin{algorithm}
    \caption{RQMC Policy Evaluation w/ Returns}
    \label{alg:rqmc_pg}
    \vspace{-0.5em}
    \begin{lstlisting}[escapeinside={(*}{*)}, language=Python]
from scipy.stats import Sobol
rqmc = Sobol((*$T \cdot \dimA$*), scramble=True)
(*$u$*) = rqmc.random((*$N$*))  # (*{\color{gray}$u \in [0, 1)^{N \times (T \cdot \dimA)}$}*)
(*$R = 0$*)
for (*$i$*) in (*$1, \dots, N$*):
   env.reset()
   for (*$t$*) in (*$1, \dots, T$*):
       (*$a^{(i)}_t$*) = (*$\mu_\params(s^{(i)}_t) + \sigma_\params(s^{(i)}_t) \odot F^{-1}(u^{(i)}_t)$*)
       (*$R$*) += env.step(*$(a^{(i)}_t)$*)
return (*$R / N$*)
    \end{lstlisting}
    \vspace{-0.5em}
\end{algorithm}

\section{LEARNING AND EVALUATION WITH RANDOMIZED QUASI-MONTE CARLO}
\label{sec:qmcpg}

We now show how to use RQMC point sets for both policy evaluation and policy improvement.
We consider both the vanilla policy gradient and actor-critic settings.
With the former, an (approximate) action-value function is not available and the policy value must be estimated by sampling trajectories from the environment and collecting returns.
This setting is more general than the actor-critic one, where the value functions are approximated with parameteric functions. The actor-critic methods tend to accelerate learning as illustrated by several recent state-of-the-art methods~\citep{fujimoto2018addressing, haarnoja2018soft}. %

\subsection{POLICY EVALUATION WITH RQMC}
\label{sec:qmc-pe}

The goal of policy evaluation is to estimate the policy value $\pvalue$ for a given policy $\policy$.

\paragraph{Through  expected returns} We estimate with
\begin{align}
   \pvalue
    &\approx \frac1{N} \sum\nolimits_{i=1}^N \sum\nolimits_{t=1}^T R(s_t^{(i)}, a_t^{(i)}),
    \label{eq:qmcpg-value} 
\end{align}
where we have $N$ trajectories $(s_1^{(i)}, a_1^{(i)}, \dots, s_T^{(i)}, a_T^{(i)})$ for all $i$.

Because the stochasticity underlying the dynamics of the environment is typically inaccessible, we use RQMC to sample the action sequence.
Concretely, we use RQMC to generate  $N$ points point $u^{(i)} \in \mathbb{R}^d$. 
Then, we obtain action $a_t^{(i)} = \policy(s_t^{(i)}, u_t^{(i)})$ by applying $\policy$ to $s_t^{(i)}$ and $u_t^{(i)}$ as in \cref{eq:reparam-trick}, where $u_t^{(i)}$ is the $t^\text{th}$ segment of length $\dimA$ from  $u^{(i)}$ (\ie, scalar entries $(t-1) \cdot \dim(A)$ to $t \cdot \dimA$ in vector $u^{(i)}$).
Pseudocode for this approach is listed in \cref{alg:rqmc_pg}.

Note that since we wish to evaluate the \emph{sum} of immediate rewards, RQMC is effectively integrating over all actions in the trajectory and each of our $N$ points ought to have dimension $d = \dimA \cdot T$ (for example, $\dimA = 4$ for a robotic arm with 4 degrees of freedom, one for each of its 4 joints).
If instead we were to use $T$ independent RQMC point sets, each with $N$ points of dimension $\dimA$, the estimator would remain unbiased w.r.t. uniformity but would lose low-discrepancy on $[0, 1)^d$.
In practice, this dependency of $d$ on both the action dimension $\dimA$ and horizon $T$ can be an impediment because RQMC implementations typically limit the integration dimension\footnote{E.g., the most popular implementation \citep{joe2008constructing} limits $d \leq 1,110$, while the latest versions of PyTorch and TensorFlow both limit $d \leq 21,201$.}.
Fortunately, this impediment can be sidestepped with learned action-value functions.

\paragraph{Through learned critic}
The above procedure can be significantly simplified if we assume access to an approximate (parametric) form of action-value function $\Q(s, a)$.
We also  assume access to a set of states used to compute the expectation over $s$ in the right-hand side of \cref{eq:qmcpg-value}.
Those states should be collected by rolling out the evaluated policy $\policy$.
Then, the expected returns can be approximated with
\begin{equation}
 \pvalue
    \approx \Exp[s_k]{ \frac1{N}\sum_{i=1}^N \hQ\left(s_k, \policy(s_k, u^{(i)}_k)\right)},
    \label{eq:qmcac-value}
\end{equation}
where $(u^{(1)}_k, \dots, u^{(N)}_k)$ is the RQMC point set over $[0, 1)^{\dimA}$ associated with state $s_k$.
Note that because $\hQ$ implicitly computes the sum of rewards over the entire trajectory, the horizon $T$ does not factor in the integration dimension $d = \dimA$, significantly reducing the sampling cost.

\subsection{POLICY ITERATION WITH RQMC}
\label{sec:qmc-pi}

In policy iteration, we compute the gradient of $\pvalue$ with respect to the  policy parameters $\params$ and use them in an iterative procedure, \eg, stochastic gradient descent, to maximize the policy value.

When $\Q$ is estimated through expected returns, policy iteration simply consists of taking the trajectories collected during policy evaluation as in~\cref{eq:qmcpg-value} and using the states $s$ and executed actions $a$ to compute the estimate in \cref{eq:rl-pg}. Thus, the RQMC procedure described above applies.

When the action-value functions are learned from a parameteric estimator $\hQ$, for example, with \cref{eq:bellman}, we can compute the gradients in two ways. The first is to use the estimator as a plug-in:
\begin{align}
    &\nabla \pvalue \nonumber\\
    &\approx \Exp[s]{\frac1{N} \sum_{i=1}^N \hQ(s, a^{(i)}) \nabla \log \policy(a^{(i)} \mid s)}
    \label{eq:plugin}
\end{align}
where $a^{(i)} = \policy(s, u^{(i)})$.

Alternatively, when $\hQ$ is differentiable with respect to the actions, for example, when $\hQ$ is instantiated as a neural network, we can use the reparameterization trick and view  the policy as a deterministic function and directly differentiate the action-value w.r.t. the policy parameters $\params$, yielding the following estimator:
\begin{align}
    &\nabla \pvalue\nonumber \\ 
    &\approx \Exp[s]{ \frac1{N} \sum_{i=1}^N \nabla_{a^{(i)}} \hQ \left(s, a^{(i)} \right) \nabla_\params \policy(s, u^{(i)})}.
    \label{eq:qmcdpg}
\end{align}
This type of deterministic policy gradient estimator~\citep{silver2014deterministic} is common in recent state-of-the-art actor-critic methods (e.g., \citet{lillicrap2015continuous, fujimoto2018addressing, haarnoja2018soft}), notably because it enables off-policy learning which drastically improves sample efficiency.
We detail and provide an implementation of soft actor-critic (SAC;~\citet{haarnoja2018soft}) with RQMC in the \supp.

In general, which approach to use is application-dependent as there are scenarios where the score-function estimator~\cref{eq:qmcac-value} yields lower-variance than the deterministic policy gradient~\cref{eq:qmcdpg} \citep[\S~3.1.2]{Gal_2016-ea}, and \emph{vice-versa} \citep[\S~8.3]{Mohamed2019-wr}.
Moreover, both estimators can be combined \citep{Gu2017-mm} but  it is unclear whether the interpolation yields any benefit.
In our experiments, we focus on~\cref{eq:qmcdpg}  as it is the current state-of-the-art method  for continuous control.

%% file: experiments.tex
\section{EXPERIMENTS}\label{sec:experiments}

In this section, we demonstrate the utility of RQMC when applied to continuous control tasks.
To that end, we propose to answer the following questions:

\begin{itemize}
    \vspace{-1em}
    \item How effective are the RQMC policy gradient and actor-critic formulations in ameliorating policy evaluation~(\S\ref{sec:exp-policy-eval}) and improvement~(\S\ref{sec:exp-policy-improv}) on a range of simulated continuous control tasks?
    \item Does RQMC  effectively reduce variance and improve gradient estimation~(\S\ref{sec:exp-analysis-gradient})?
    \item How does RQMC fare against other variance reduction techniques, and can it complement them~(\S\ref{sVRT})? 
    \vspace{-1em}
\end{itemize}

\paragraph{Tasks} We conduct our study on three sets of standardized tasks: Brownian motion, Linear-Quadratic Regulator (LQR), as well as the popular \mujoco robotics suite~\citep{todorov2012mujoco}.
Throughout, we estimate $\Q$ by sampling trajectories (\cref{eq:qmcpg-value}) for the Brownian motion and LQR tasks, and learn a parameterized $\hQ(s, a)$ with the soft actor-critic (SAC) of~\citet{haarnoja2018soft} for the \mujoco tasks.
We randomize point sets with the left matrix scramble and digital shift for Brownian motion and LQR, and with Owen's scrambling~\citep{Owen1998-ad} for \mujoco tasks.

\input{results/fig_policy_evaluation.tex}

The Brownian motion environment -- as studied in \cite[\S~4.5]{LEcuyer2008-tm} -- consists of a 1D point-mass particle whose objective is to reach the origin.
At every step, an action is sampled from a Gaussian distribution parameterized by its mean and covariance.
The reward is the Euclidean distance of the particle's current position to the origin, and we set the horizon to 20 timesteps.

The LQR environment is a well-studied testbed from the optimal control literature which provides a unified formalism for many continuous control problems.
LQR has also been the focus of recent theoretical studies of policy gradient methods \citep{fazel2018global, recht2019tour}.
The transition dynamics of LQR are a linear function of the state and action, while the reward is a quadratic function of both values.
In our experiments, we randomly sample transition and reward matrices, with an 8-dimensional state-space, 6-dimensional action-space, and horizon set to 20.
As is standard in the literature, we use a linear Gaussian policy.
For a more thorough treatment on LQR, we refer the reader to \citep{10.5555/578807}.

We use the \mujoco robotic environments available in the standardized OpenAI Gym suite~\citep{brockman2016openai}.
For those experiments, we use a tanh-Gaussian policy whose mean and diagonal covariance are given by a 2-layer neural network.
Unless specified otherwise, we use Adam~\citep{kingma2014adam} as the optimization algorithm, share the learning rate among MC and RQMC, and set the horizon to 1000 timesteps.
The main text presents results on \walker, \cheetah, and \ant, with results on \hopper and \swimmer in the \supp.

\subsection{POLICY EVALUATION}\label{sec:exp-policy-eval}

Our first set of experiments compares the sample efficiency of MC and RQMC.
To do so, we take a fixed policy (random policy in Brownian motion, optimal policy for LQR and trained policy for \mujoco) and sample a predefined number of trajectories using either MC or RQMC.
With those trajectories, we compute an estimated $\hat{V}^\pi$ of the expected return of the policy and compare it against the ground-truth policy value $V^\pi$.
For the Brownian motion and LQR, the ground-truth is computed analytically and the estimation $\hat{V}^\pi$ is computed with \cref{eq:qmcpg-value} with actions generated using MC or RQMC.

For the \mujoco tasks, the ground-truth $V^\pi$ is estimated on $2^{10}$ states for each of which we sample $2^{16}$ actions with MC.
The estimated $\hat{V}^\pi = \Exp[s]{\frac1{N} \sum_{i=1}^N \hQ(s, \policy(s, u^{(i)}))}$ uses a $\hQ$-function learned with SAC, and $u^{(i)}$ are sampled with either RQMC or MC.
By varying the number of samples to estimate $\hat{V}^\pi$ we can observe the convergence rate of different integration methods.
For each method, we repeat those experiments 30 times and report the mean squared error $\vert V^\pi - \hat{V}^\pi\vert^2$ in \cref{fig:exp-pe-main}.

For the Brownian motion, RQMC outperforms MC when using more than 128 trajectories.
On LQR, RQMC outperforms MC with as little as 4 trajectories.
Importantly, RQMC converges significantly faster than MC and ultimately reduces the evaluation error by more than an order of magnitude on both of these tasks.

For \mujoco, RQMC significantly outperforms MC when using as little as 16 actions per state.
The gap between MC and RQMC continues to increase with more actions, and RQMC reaches over an order of magnitude lower estimation error with 2048 actions per states on all tasks.

\subsection{POLICY IMPROVEMENT}\label{sec:exp-policy-improv}

\input{results/fig_policy_improvement.tex}

To examine RQMC's effectiveness in learning a policy, we use the LQR and \mujoco testbeds to optimize a randomly initialized policy.
On LQR we train with vanilla policy gradient (VPG) using \cref{eq:qmcac-value} for 10M environment interactions and compute the gradients using 16 trajectories drawn with either MC or RQMC.
On \mujoco, we train with SAC for 3M environment interactions and draw 8 actions per state using MC or RQMC during off-policy learning.
For reference, we also include learning curves for DDPG~\citep{lillicrap2015continuous} and TD3~\citep{fujimoto2018addressing}.
At test-time, we do not sample actions and instead use the mean of the policy.
We use identical learning hyper-parameters for all methods (chosen to maximize convergence rate of MC); details are reported in the \supp.

\cref{fig:exp-pi-main} shows the mean convergence curves and standard errors computed over 15 different random seeds.
On LQR, RQMC converges approximately 4 times faster than MC and also reaches lower asymptotic cost.
On \mujoco, RQMC outperforms MC in all cases and especially later in training where noisy gradients hinder exploitation.
Notably, swapping MC for RQMC yields improvements comparable to those obtained when upgrading the learning algorithm from TD3 to SAC.

\subsection{IMPROVED GRADIENT ESTIMATION}\label{sec:exp-analysis-gradient}

\input{results/fig_gradient_variance.tex}

In this section, we verify the hypothesis that RQMC improves learning by reducing the variance of the stochastic gradient.
To demonstrate this empirically, we take a trained policy and compare the variance and alignment of its gradient when estimated with MC versus RQMC.
On LQR, we use the optimal policy obtained analytically and estimate gradients by replacing $\Q$ with the sum of rewards in \cref{eq:rl-pg}; on \mujoco, we train a policy until convergence with SAC and use \cref{eq:qmcdpg} to compute gradients.
We define gradient variance be the trace of the covariance matrix $\Exp[\hat{g}]{(\hat{g} - g)^\top (\hat{g} - g)}$, where $\hat{g}$ is the gradient estimated with a fixed number of trajectories (for LQR) or actions (for \mujoco), and $g$ is the ground-truth gradient computed with $48,000$ trajectories (for LQR) or $2^{16}$ actions (for \mujoco).
Similarly, we define gradient alignment $1 - \cos(\hat{g}, g)$, where $\cos(\hat{g}, g)$ is the cosine similarity between estimated and ground-truth gradient.
We compute 95\% confidence intervals by repeating each measurement with 30 different random seeds.

\input{results/fig_gradient_alignment.tex}

\cref{fig:exp-grad-var-main} reports the gradient variance for MC and RQMC.
On LQR, we observe that RQMC always gives lower gradient variance regardless of the number of trajectories.
On \mujoco, we observe similar results where RQMC compares favorably with as little as 4 actions, and converges at a faster rate than MC.

We observe similar results for gradient alignment in \cref{fig:exp-grad-cos-main}.
On all settings, RQMC provides gradient estimates that converge faster and are better aligned with the ground-truth than MC.

\subsection{COMBINATION WITH VARIANCE REDUCTION TECHNIQUES}
\label{sVRT}

\input{results/fig_vrt_action_v_dyna.tex}

In our last experiments, we compare RQMC against other variance reduction techniques and test whether it can complement those existing approaches.
We focus on policy improvement with VPG on LQR since SAC already includes several advances that address issues related to variance reduction (\eg, deterministic policy gradient, Adam, off-policy learning).
We consider two alternative variance reduction techniques: control variates (CV) and variance-reduced optimizers.
For control variates, we use the generalized advantage estimator~\citep{schulman2015high} and learn a linear baseline $b(s)$ to estimate the returns $V^\policy$ when starting from state $s$.
This technique is ubiquituous in continuous control for actor-critic methods that do not directly learn an action-value function $\Q$~\citep{mnih2016asynchronous, schulman2015trust}.
For variance-reduced optimizers, we compare SGD (with momentum~\citep{Polyak1964-pu}) to Accelerated SGD (ASGD;~\citet{kidambi2018insufficiency,jain2018accelerating}), one of the first methods with provable accelerated asymptotic convergence with constant learning rate on noisy linear regression problems.

We compare MC, CV, ASGD, and RQMC in \cref{fig:exp-vrts-main} (left panel) and observe that all variance reduction techniques outperform MC, with ASGD and RQMC performing best.
When combining those techniques \cref{fig:exp-vrts-main} (right panel), we see that adding CV and ASGD to MC (purple-dotted curve) performs on par with ASGD alone, suggesting that the benefits of CV are already captured by ASGD.
On the other hand, adding CV and ASGD on top of RQMC (black-starred curve) accelerates convergence, thus demonstrating that RQMC complements these variance reduction techniques.

%% file: results/fig_policy_evaluation.tex
\begin{figure*}[t]
    \vspace{-1.0em}
    \begin{center}
        \includegraphics[width=0.19\linewidth]{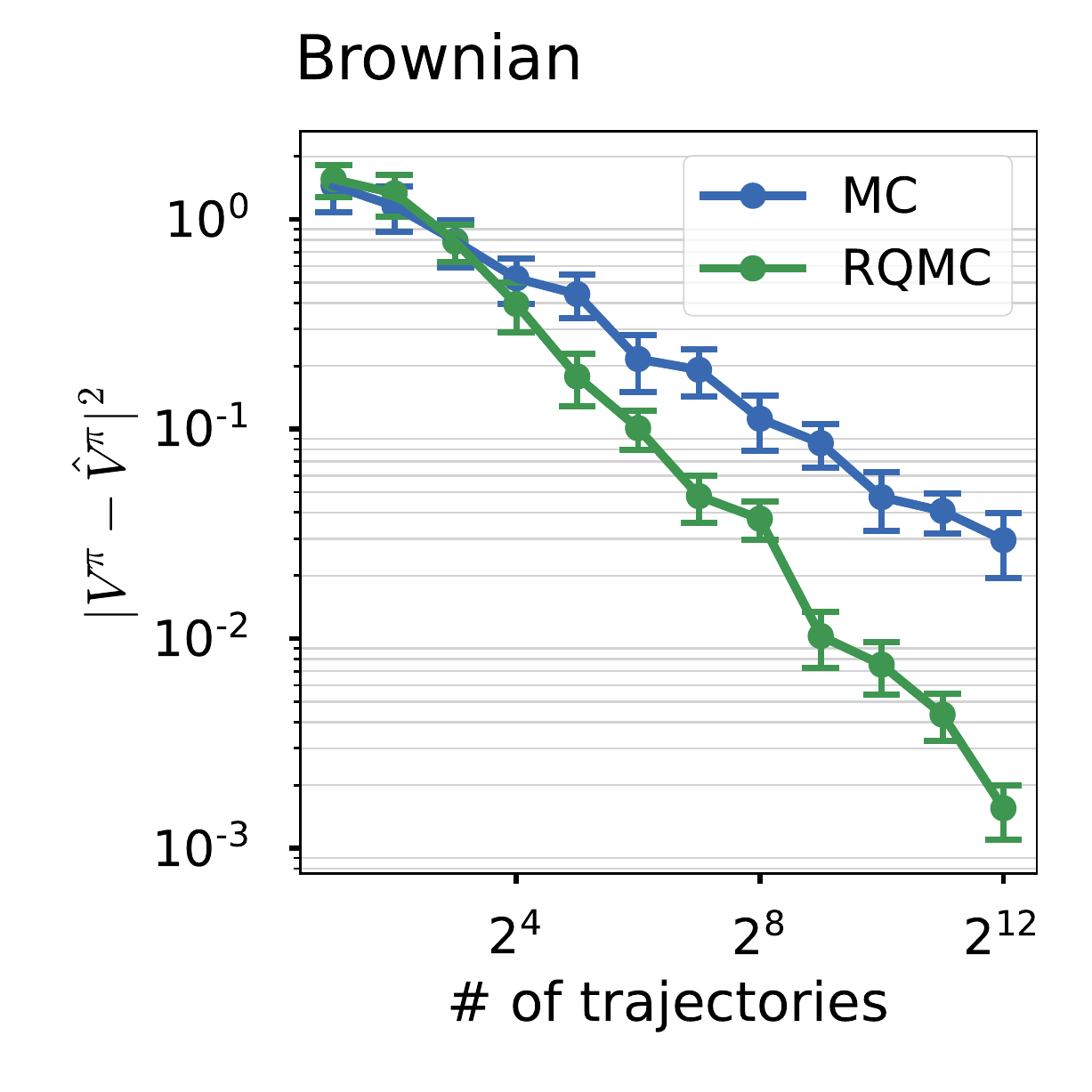}
        \includegraphics[width=0.19\linewidth]{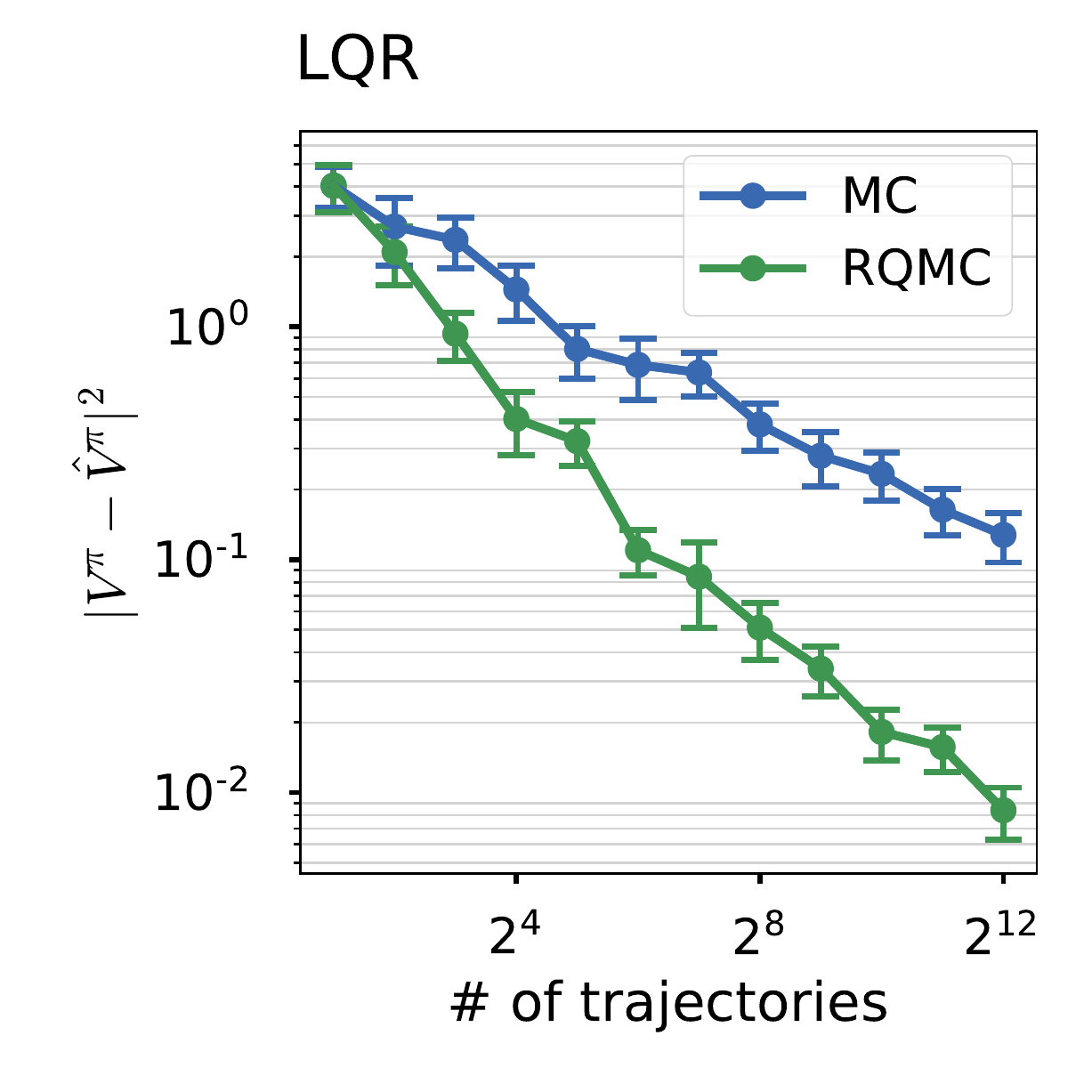}
        \includegraphics[width=0.19\linewidth]{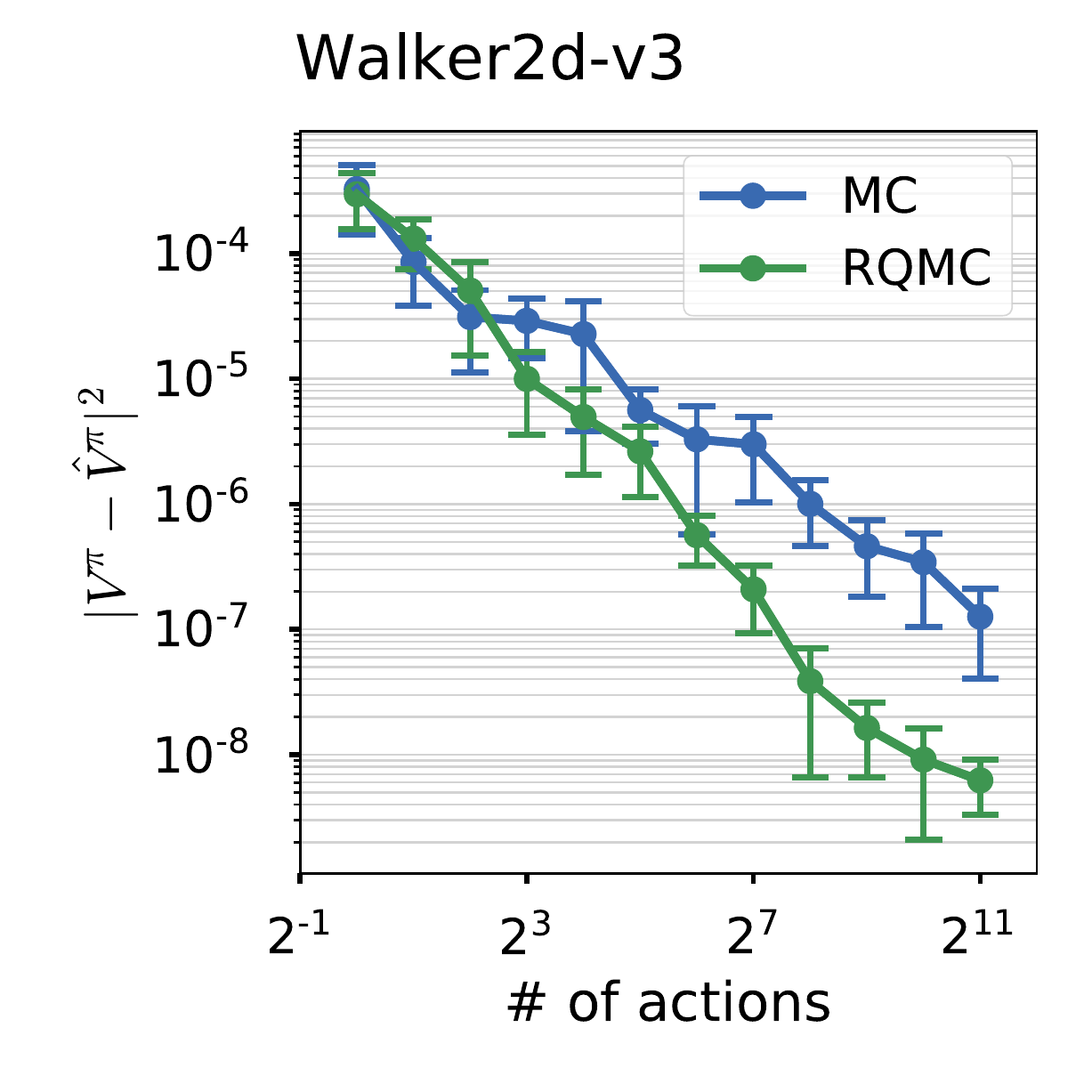}
        \includegraphics[width=0.19\linewidth]{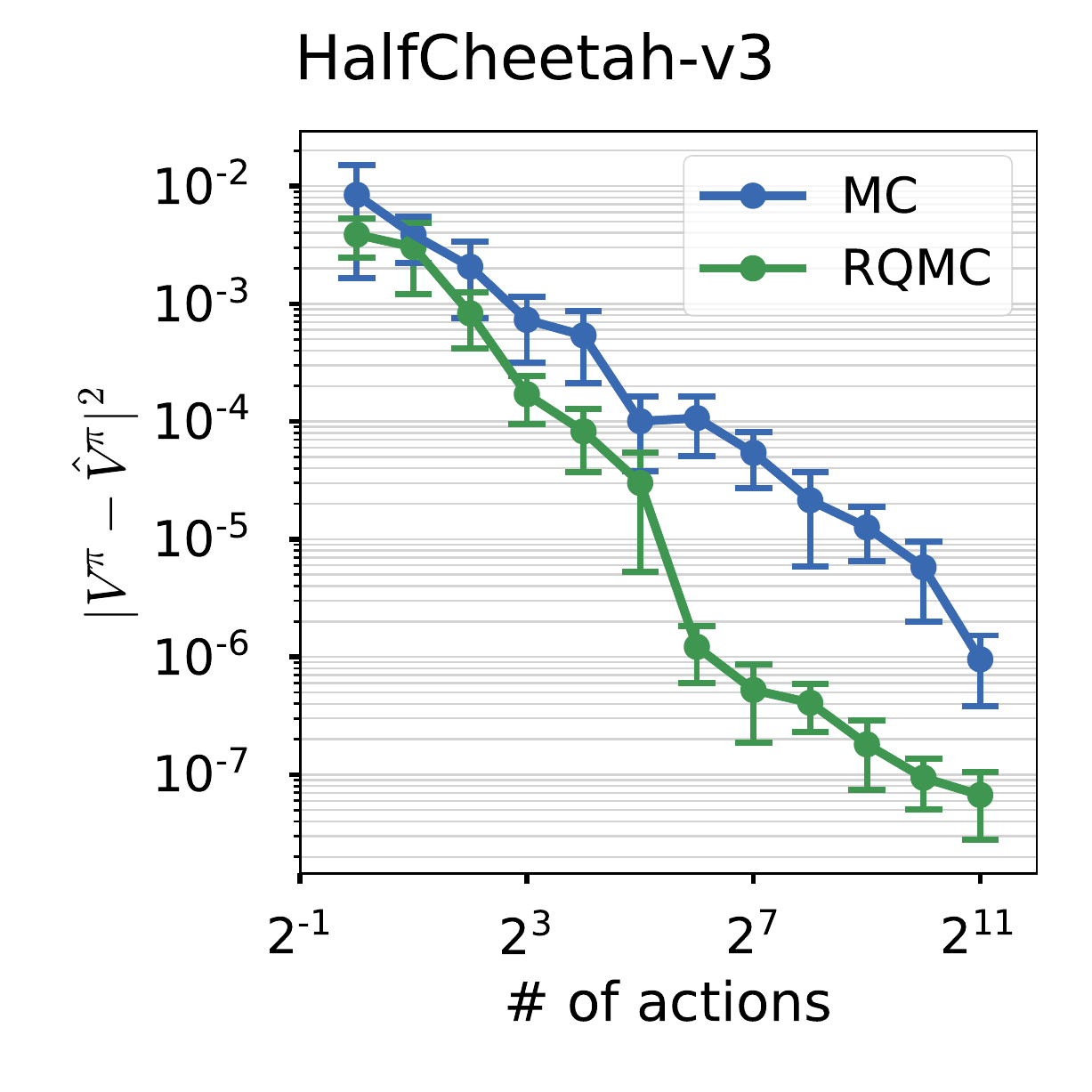}
        \includegraphics[width=0.19\linewidth]{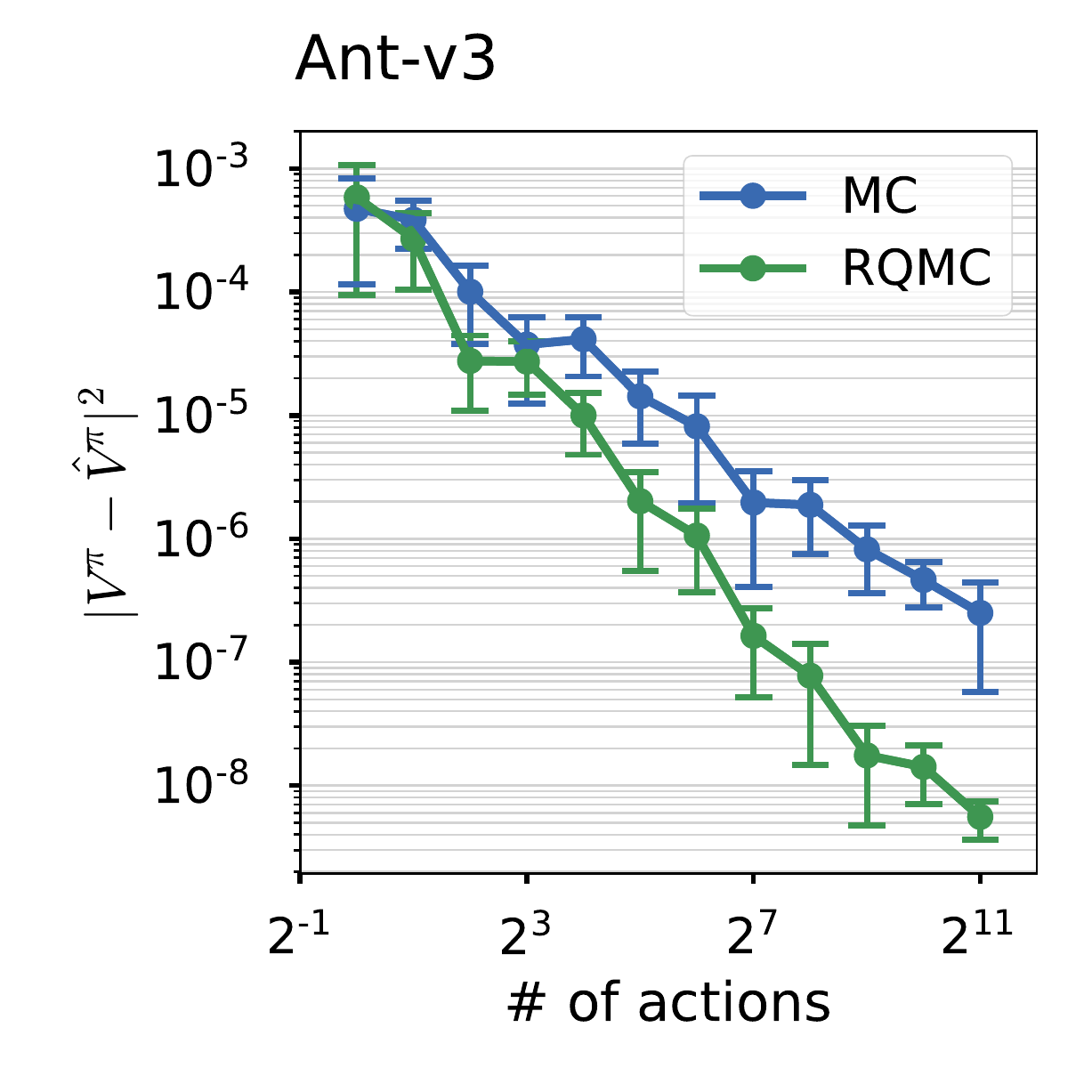}
    \end{center}
    \caption{\small 
        \textbf{RQMC reduces value estimation error.}
        RQMC is  much more efficient than MC in estimating the value of a policy on a given number of samples.
        As suggested by the theory, the gap between RQMC and MC grows with the number of trajectories (in Brownian and LQR) or the number of sampled actions per state (in \mujoco tasks, using the learned $\hQ$).
        For \mujoco tasks where ground-truth gradient is not available, we approximate it using $2^{16}$ actions.
    }
    \label{fig:exp-pe-main}
    \vspace{-1.0em}
\end{figure*}

%% file: results/fig_policy_improvement.tex
\begin{figure}
    \vspace{-1.0em}
    \begin{center}
        \includegraphics[width=0.49\linewidth]{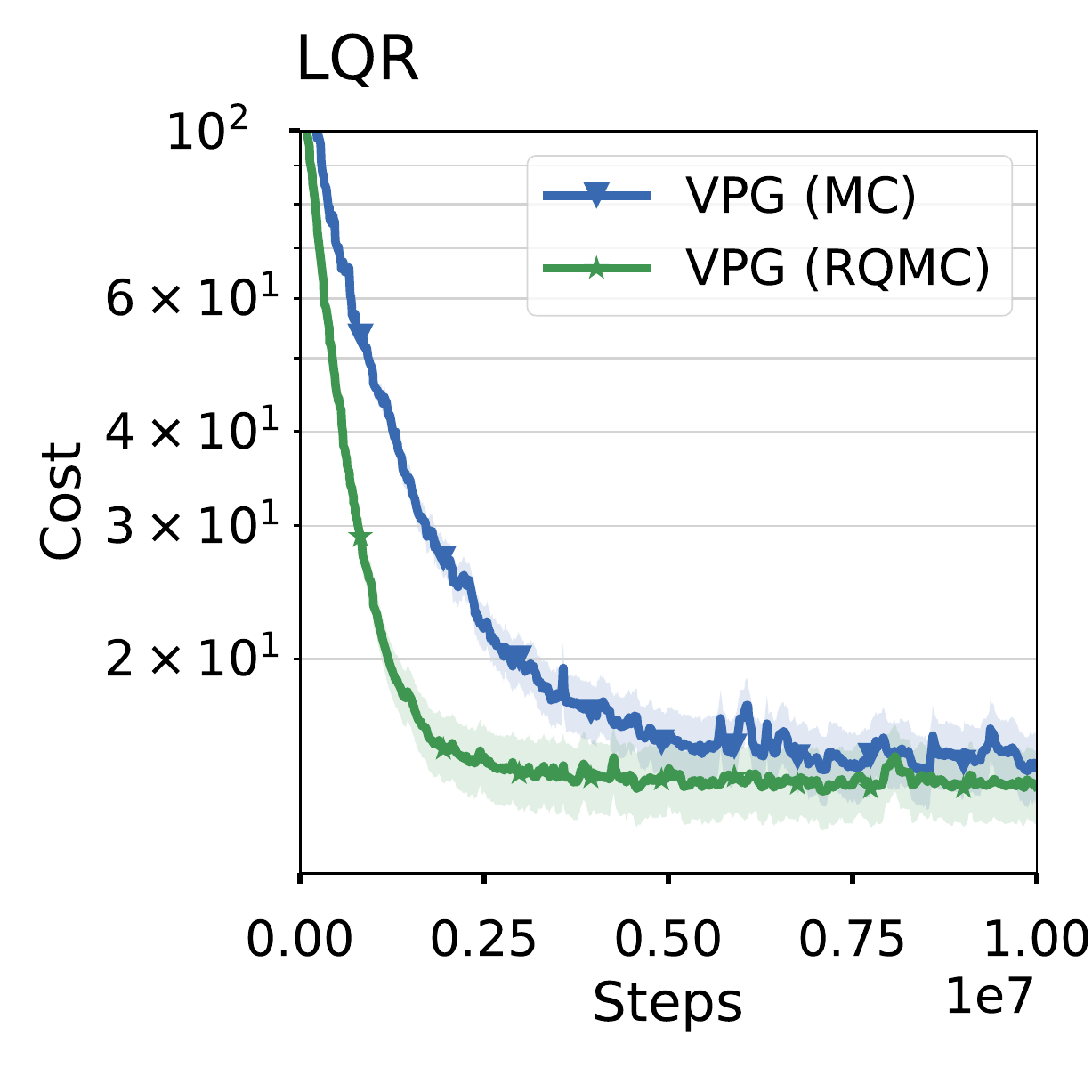}
        \includegraphics[width=0.49\linewidth]{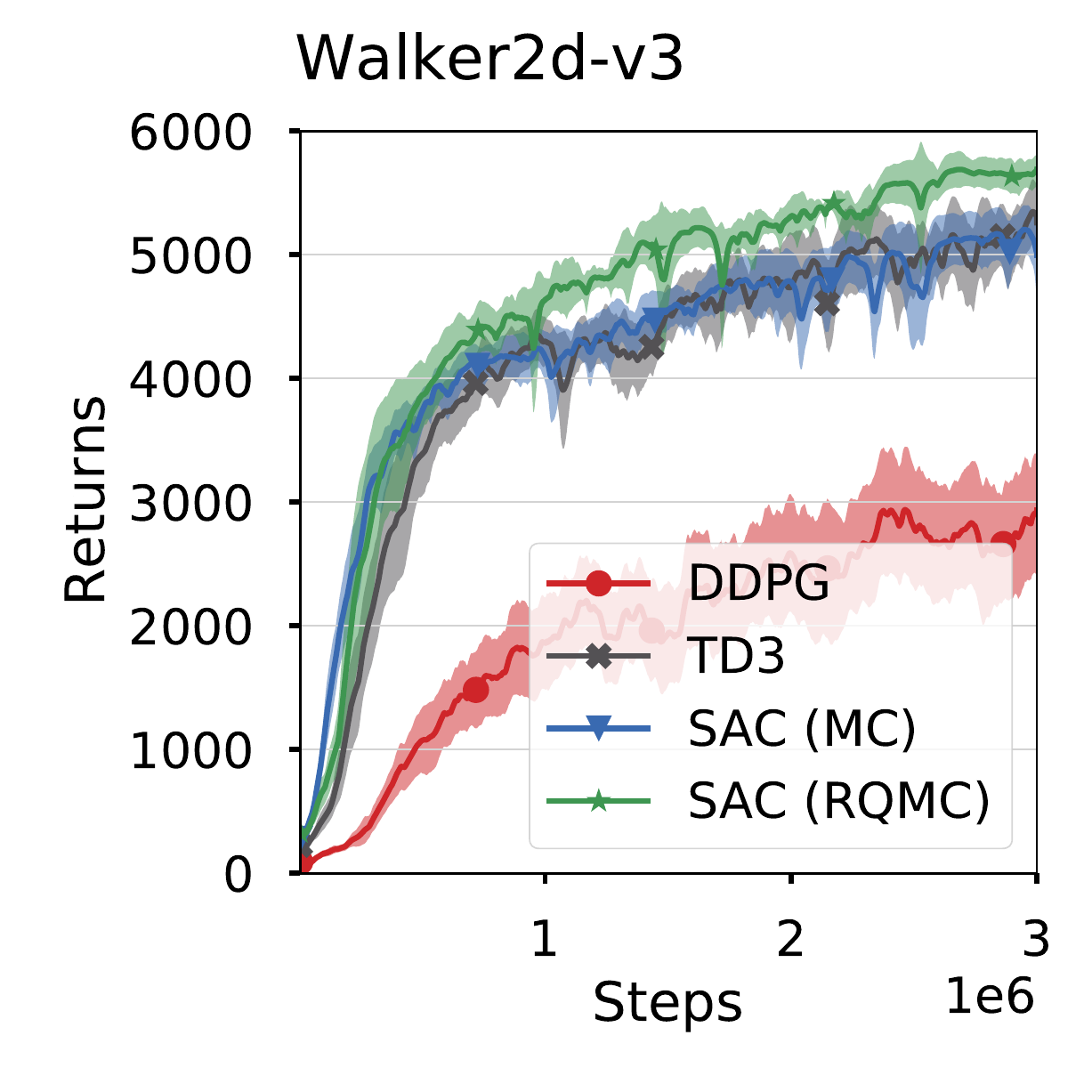} \\
        \includegraphics[width=0.49\linewidth]{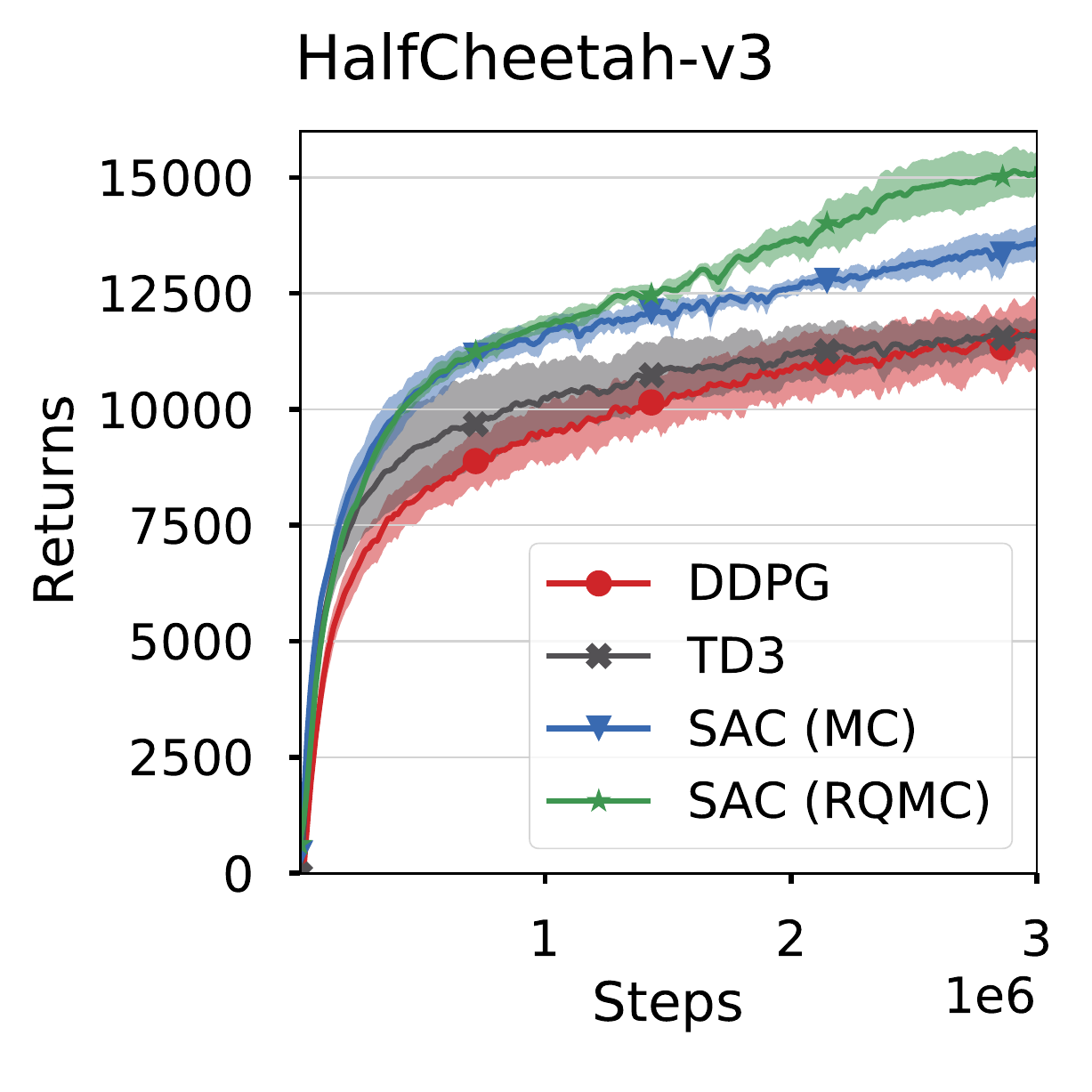}
        \includegraphics[width=0.49\linewidth]{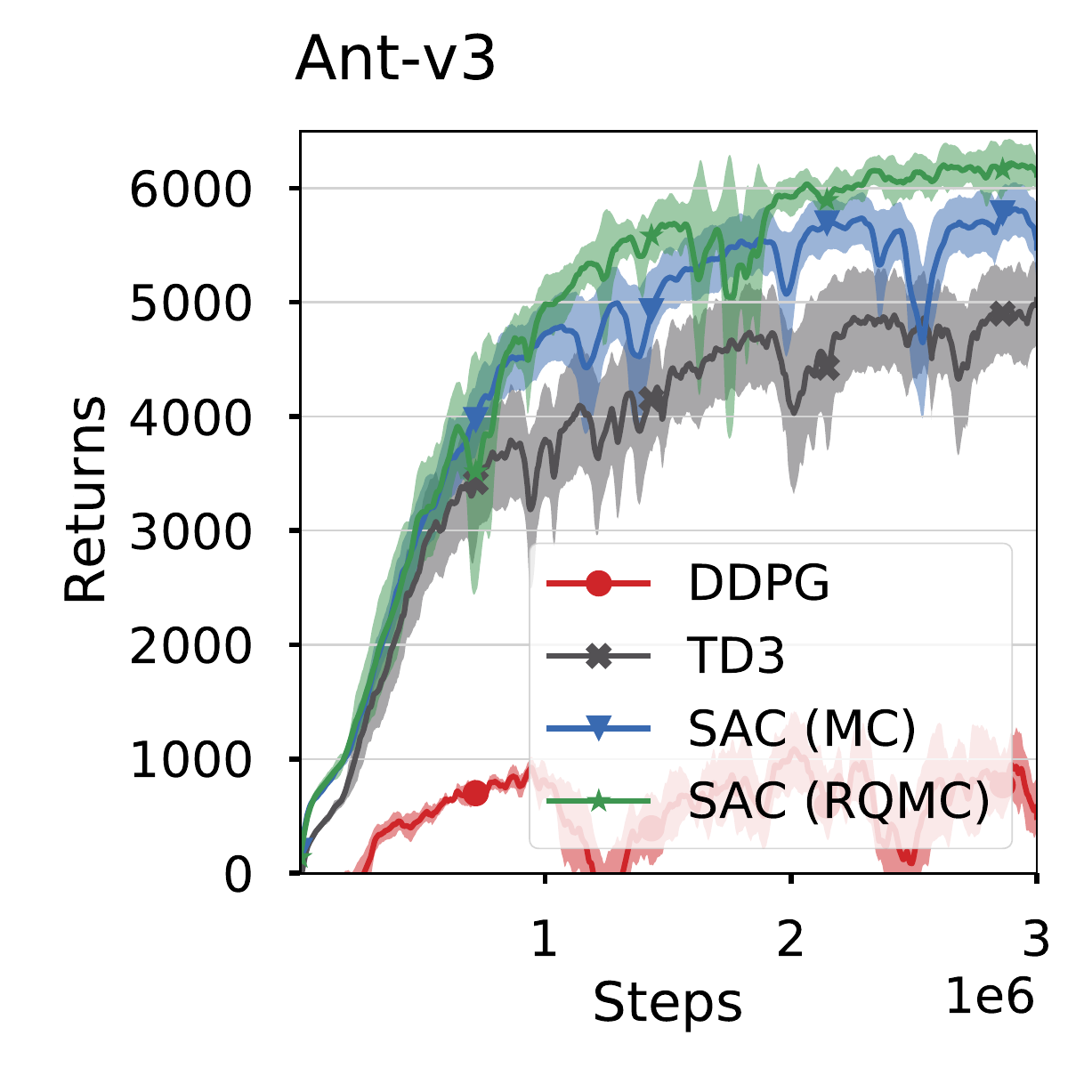}
        \caption{\small 
            \textbf{RQMC improves policy learning.}
            Using RQMC for policy learning outperforms MC on LQR and Mujoco tasks.
            In particular, RQMC improves upon MC with SAC -- a state-of-the-art actor-critic method -- in terms asymptotic performance on all Mujoco tasks.
        }
        \label{fig:exp-pi-main}
    \end{center}
    \vspace{-2.0em}
\end{figure}

%% file: results/fig_gradient_variance.tex
\begin{figure}
    \vspace{-1.0em}
    \begin{center}
        \includegraphics[width=0.49\linewidth]{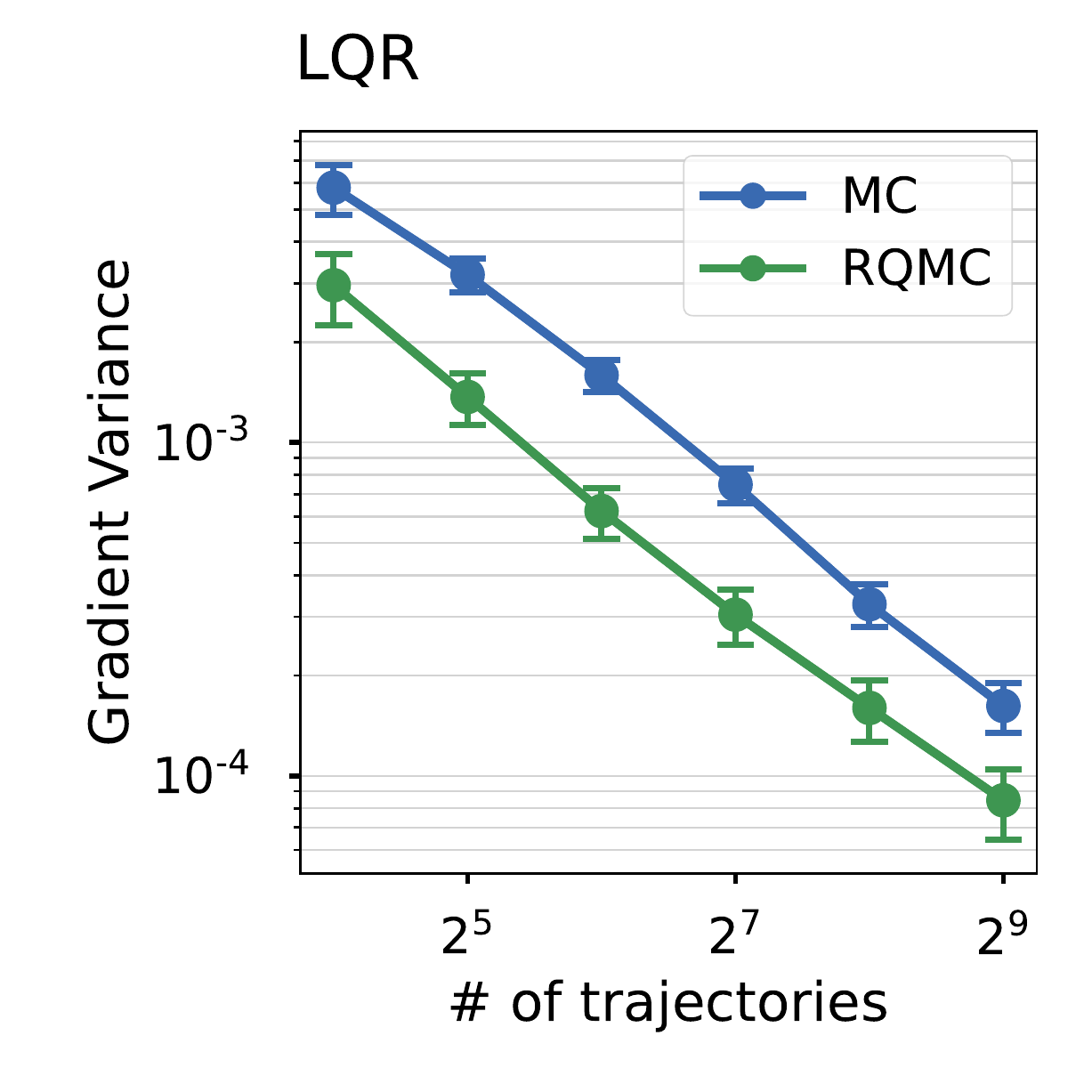}
        \includegraphics[width=0.49\linewidth]{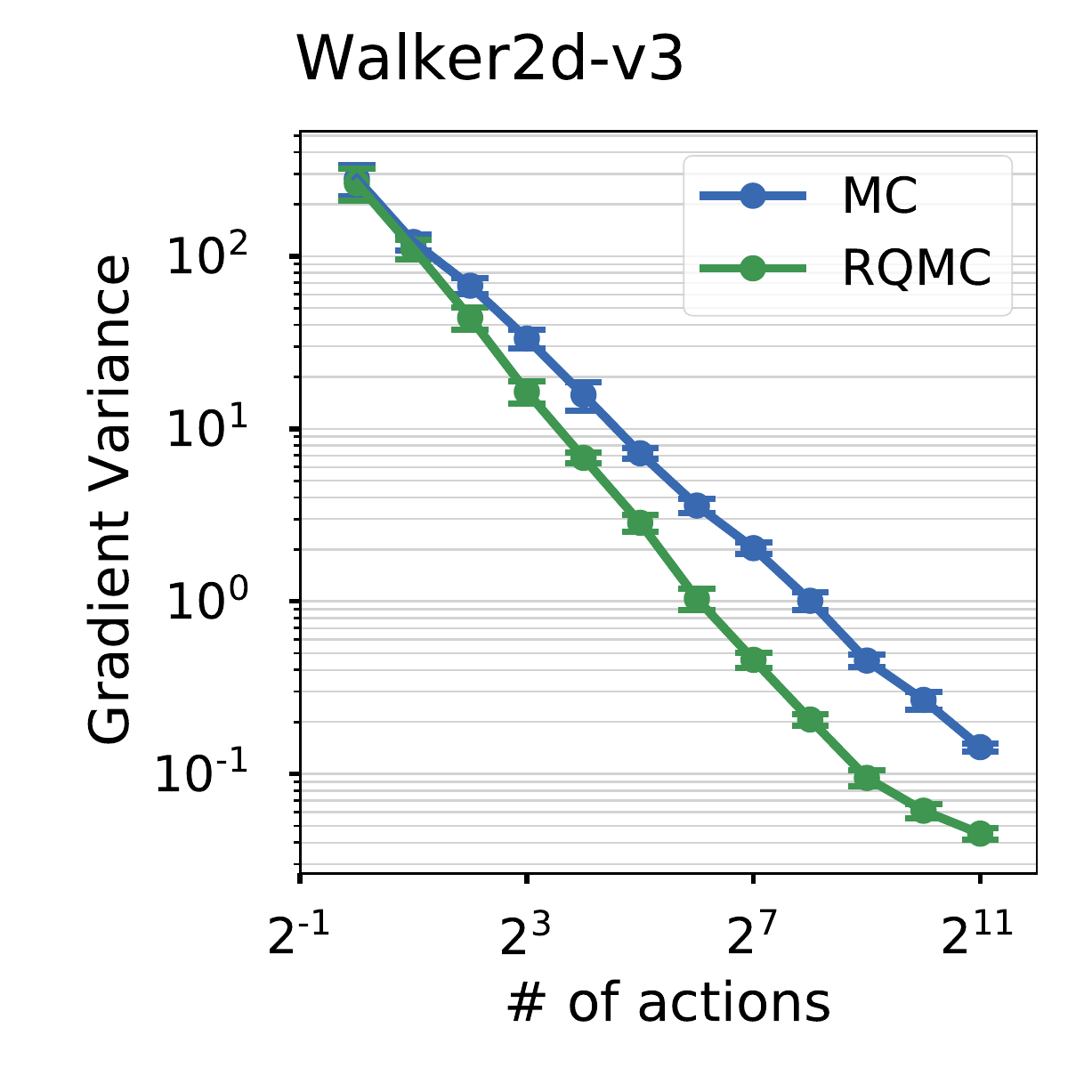} \\
        \includegraphics[width=0.49\linewidth]{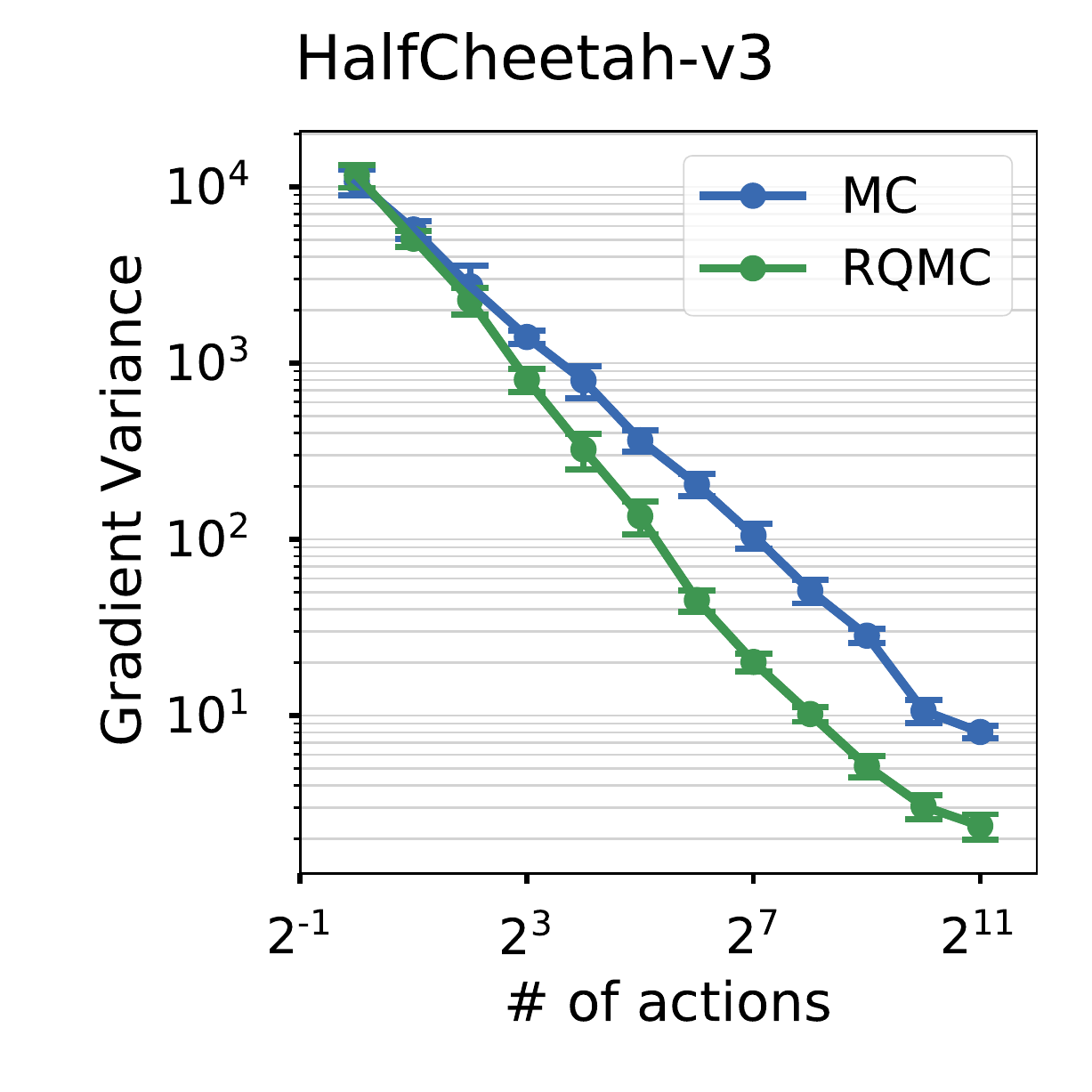}
        \includegraphics[width=0.49\linewidth]{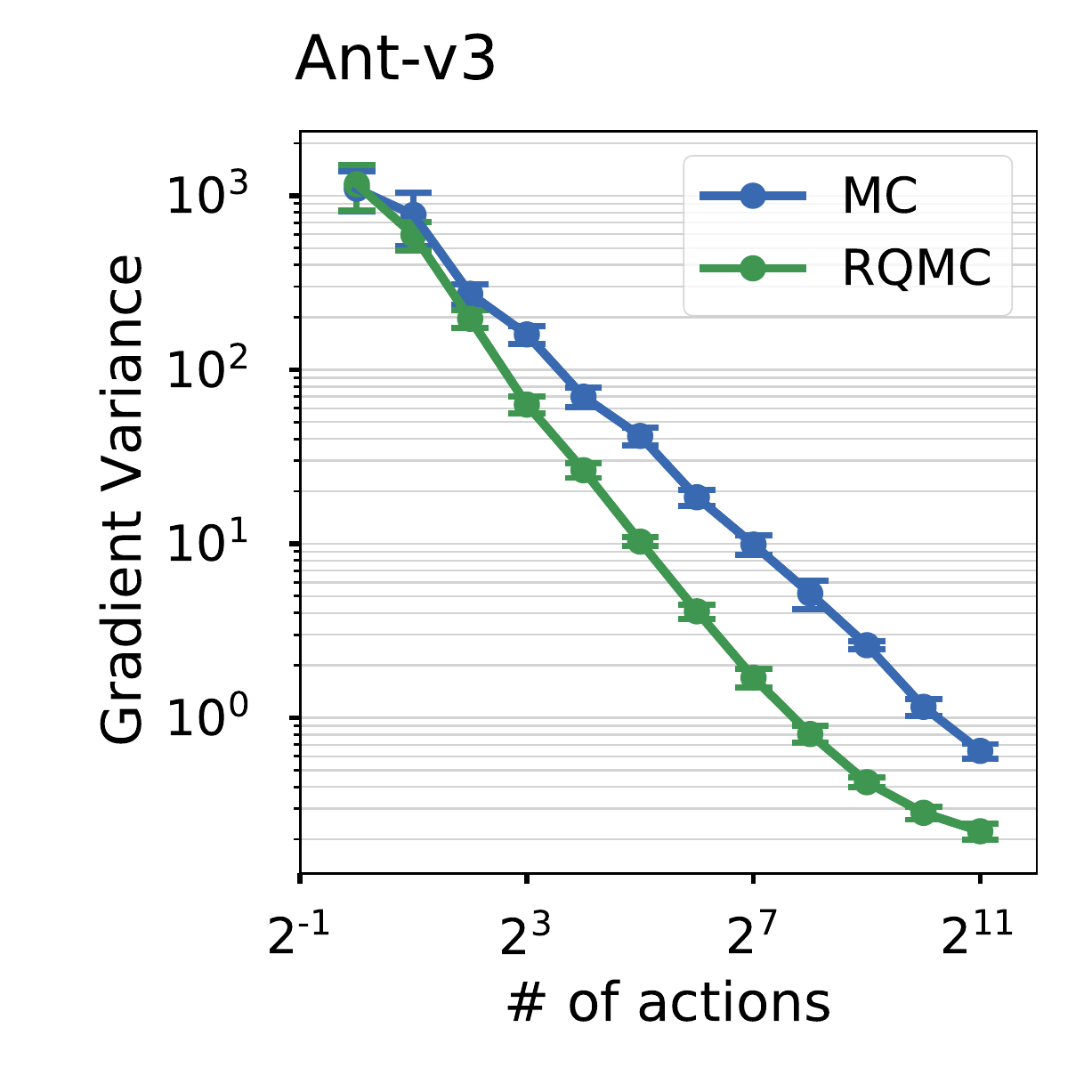}
    \end{center}
    \caption{\small 
        \textbf{RQMC reduces gradient variance.}
        On both LQR and \mujoco tasks, RQMC achieves lower gradient variance than MC for the same number of trajectories.
        Here, variance refers to the trace of the gradient covariance matrix.
        The $95\%$ confidence intervals are computed over 30 random seeds.
    }
    \label{fig:exp-grad-var-main}
    \vspace{-1.0em}
\end{figure}

%% file: results/fig_gradient_alignment.tex
\begin{figure}
    \vspace{-1.0em}
    \begin{center}
        \includegraphics[width=0.49\linewidth]{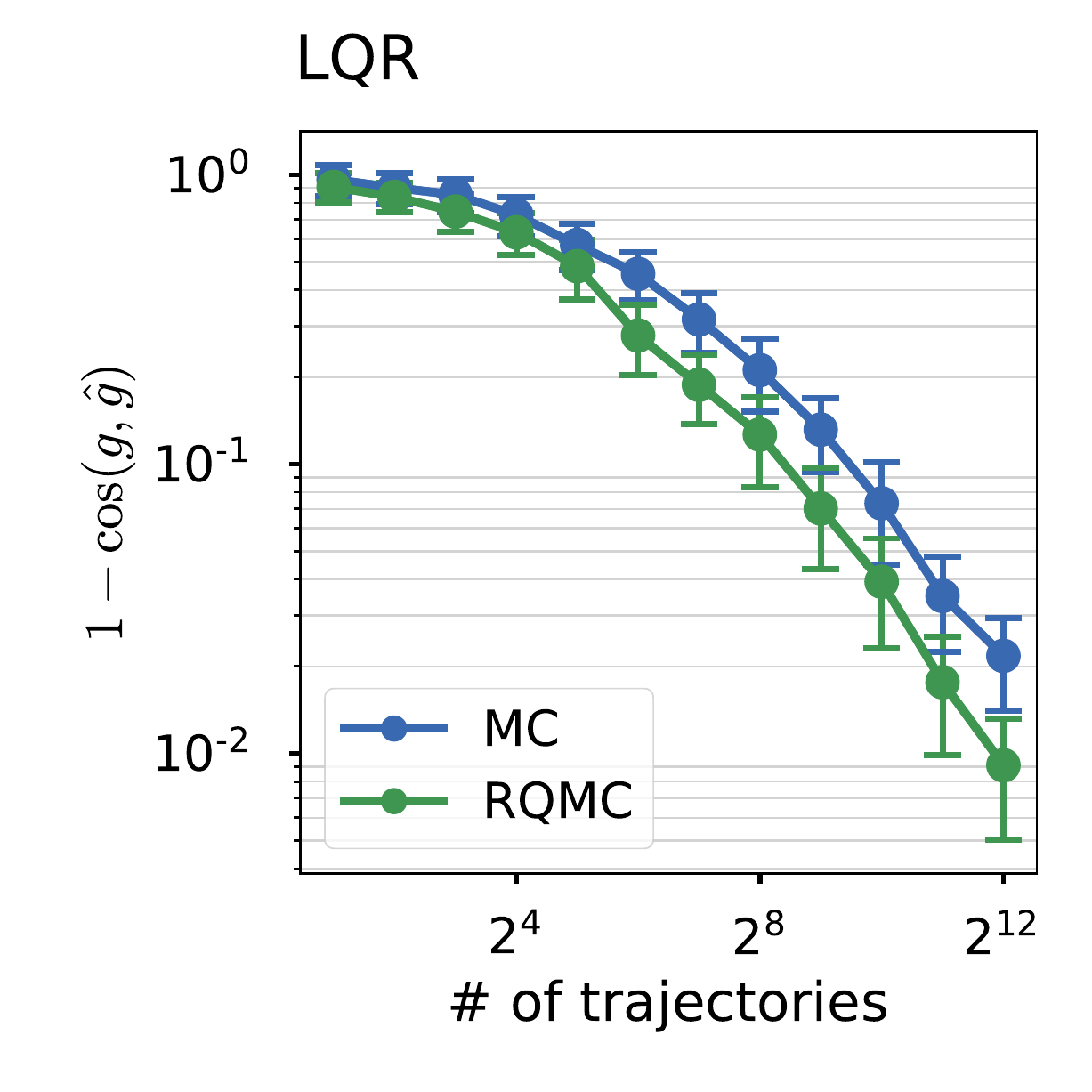}
        \includegraphics[width=0.49\linewidth]{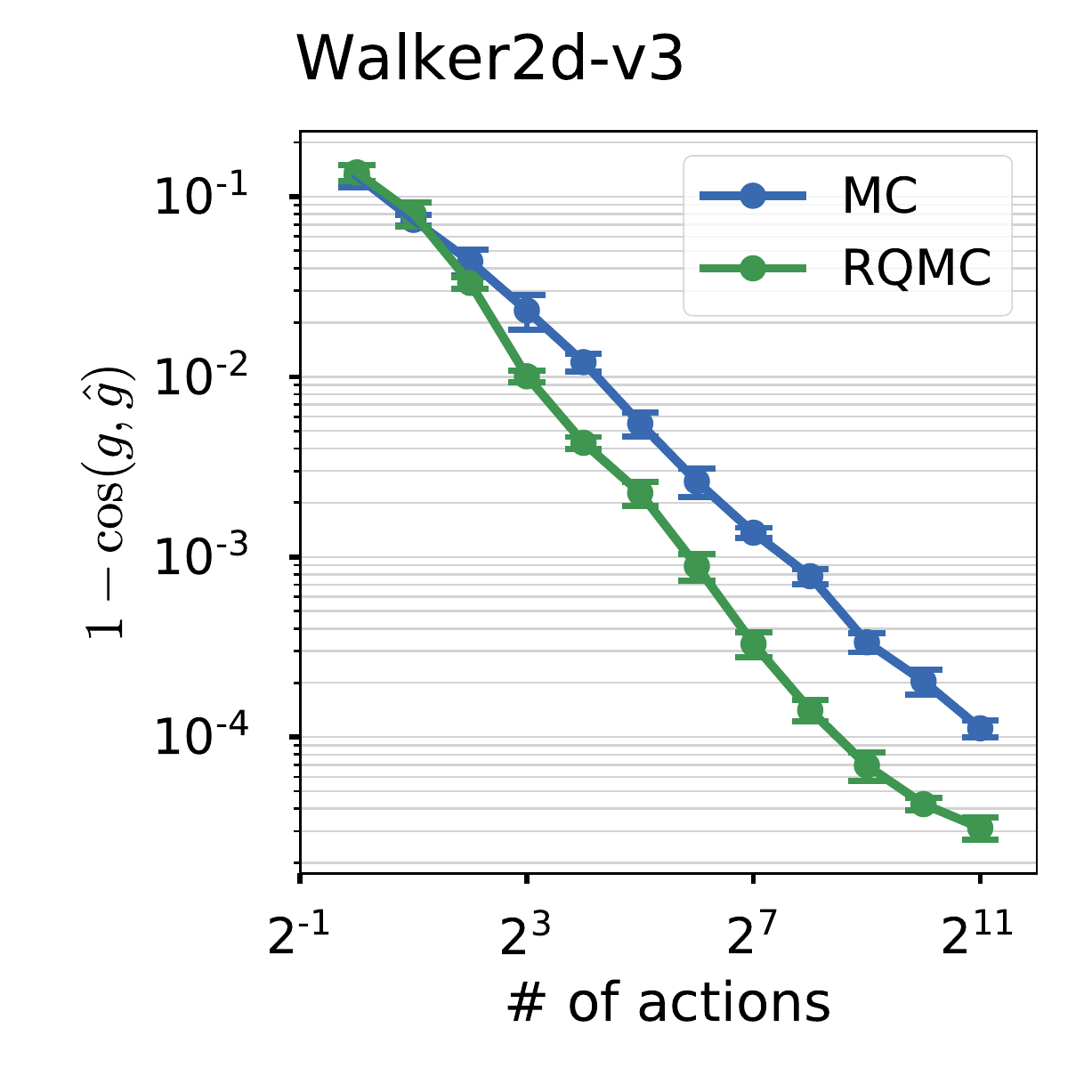} \\
        \includegraphics[width=0.49\linewidth]{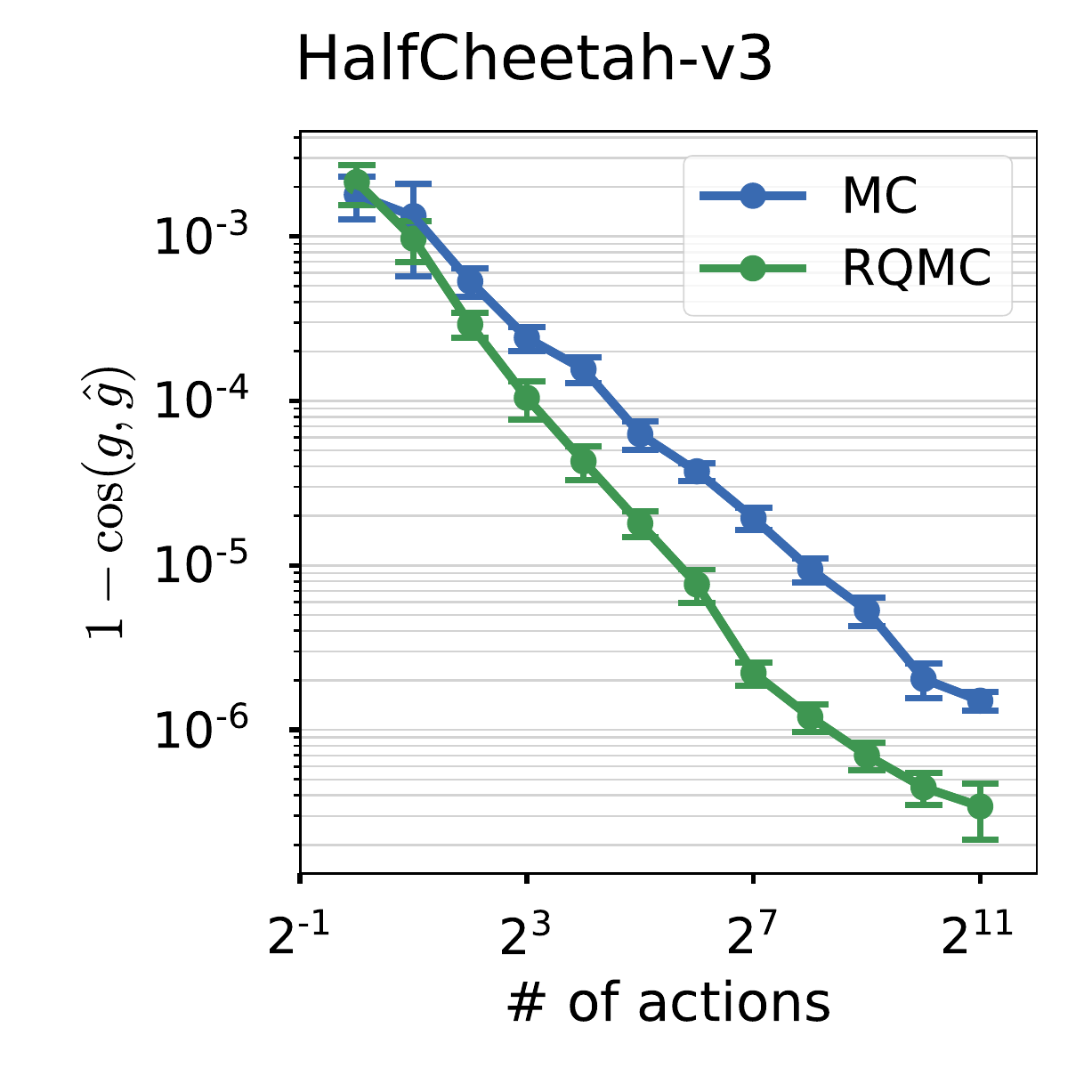}
        \includegraphics[width=0.49\linewidth]{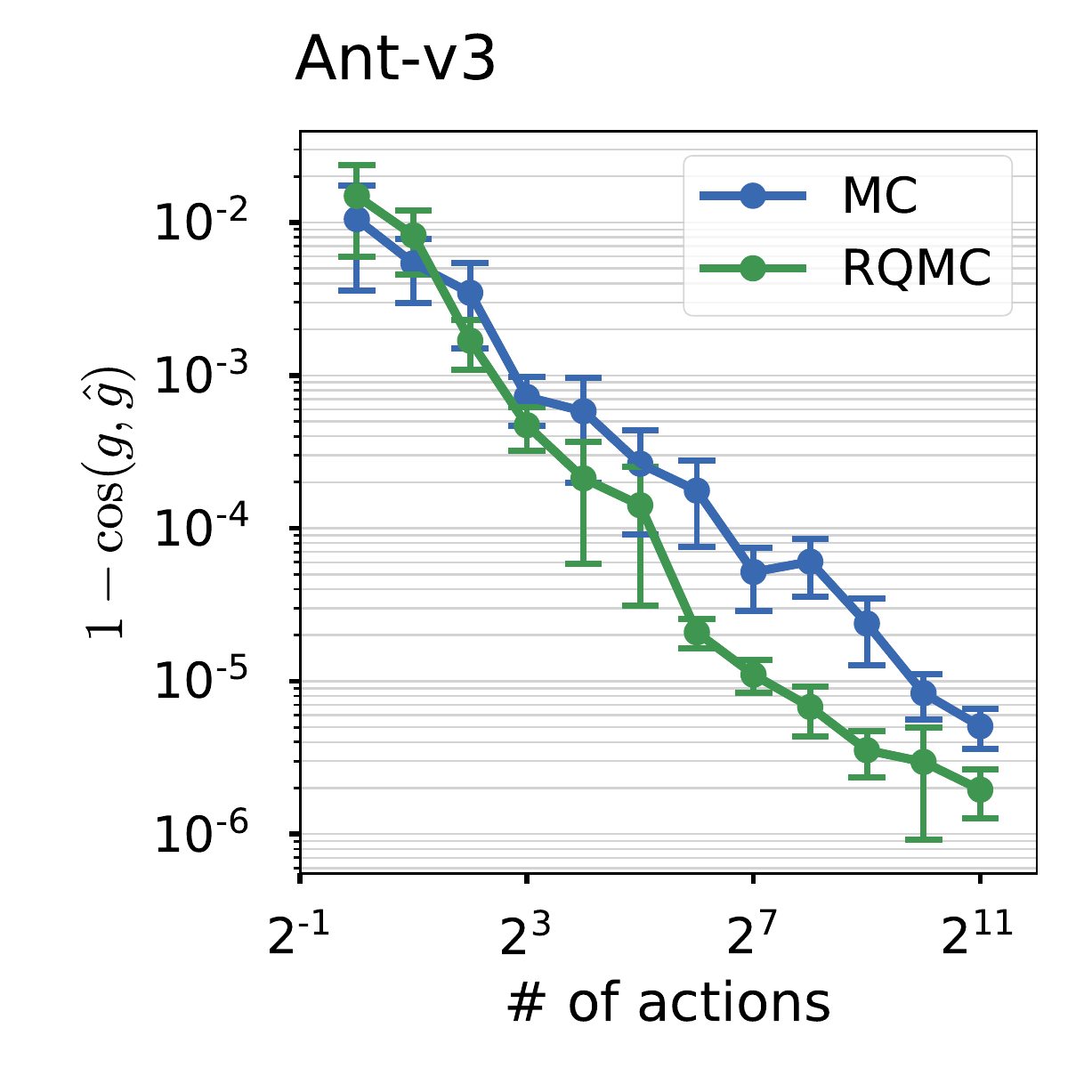}
    \end{center}
    \caption{\small 
        \textbf{RQMC improves gradient alignment.}
        For a given number of trajectories, the gradient direction computed with RQMC is better aligned than when computed with MC.
        The $y$-axis displays the angle between ground-truth and stochastic gradient.
        On LQR, the ground-truth is computed with $48k$ trajectories; on \mujoco, it is estimated using $2^{16}$ actions.
        The $95\%$ confidence intervals are computed over 30 random seeds.
    }
    \label{fig:exp-grad-cos-main}
    \vspace{-1.0em}
\end{figure}

%% file: results/fig_vrt_action_v_dyna.tex
\begin{figure}
    \vspace{-1.0em}
    \begin{center}
        \includegraphics[width=0.42\linewidth]{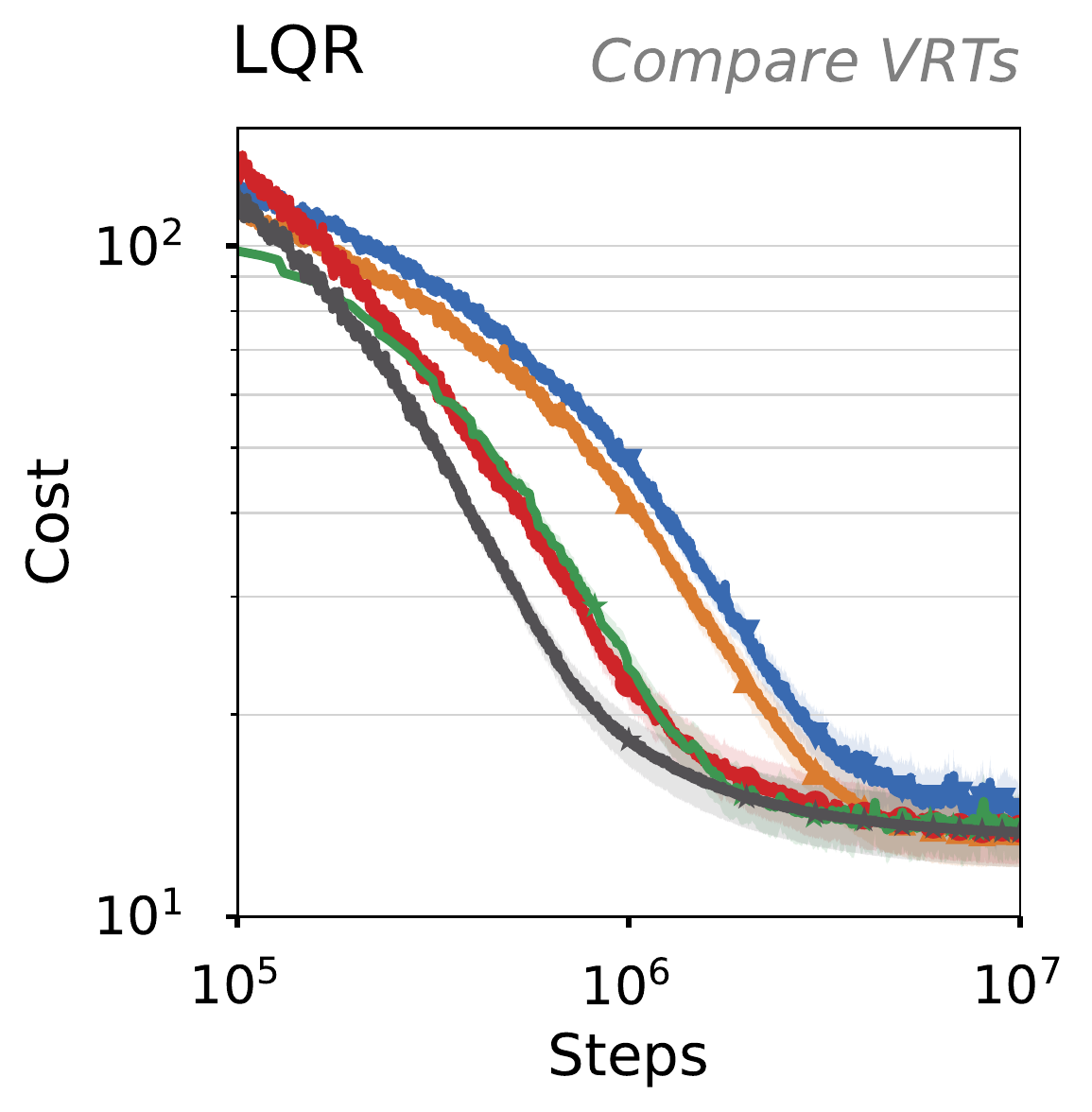}
        \includegraphics[width=0.42\linewidth]{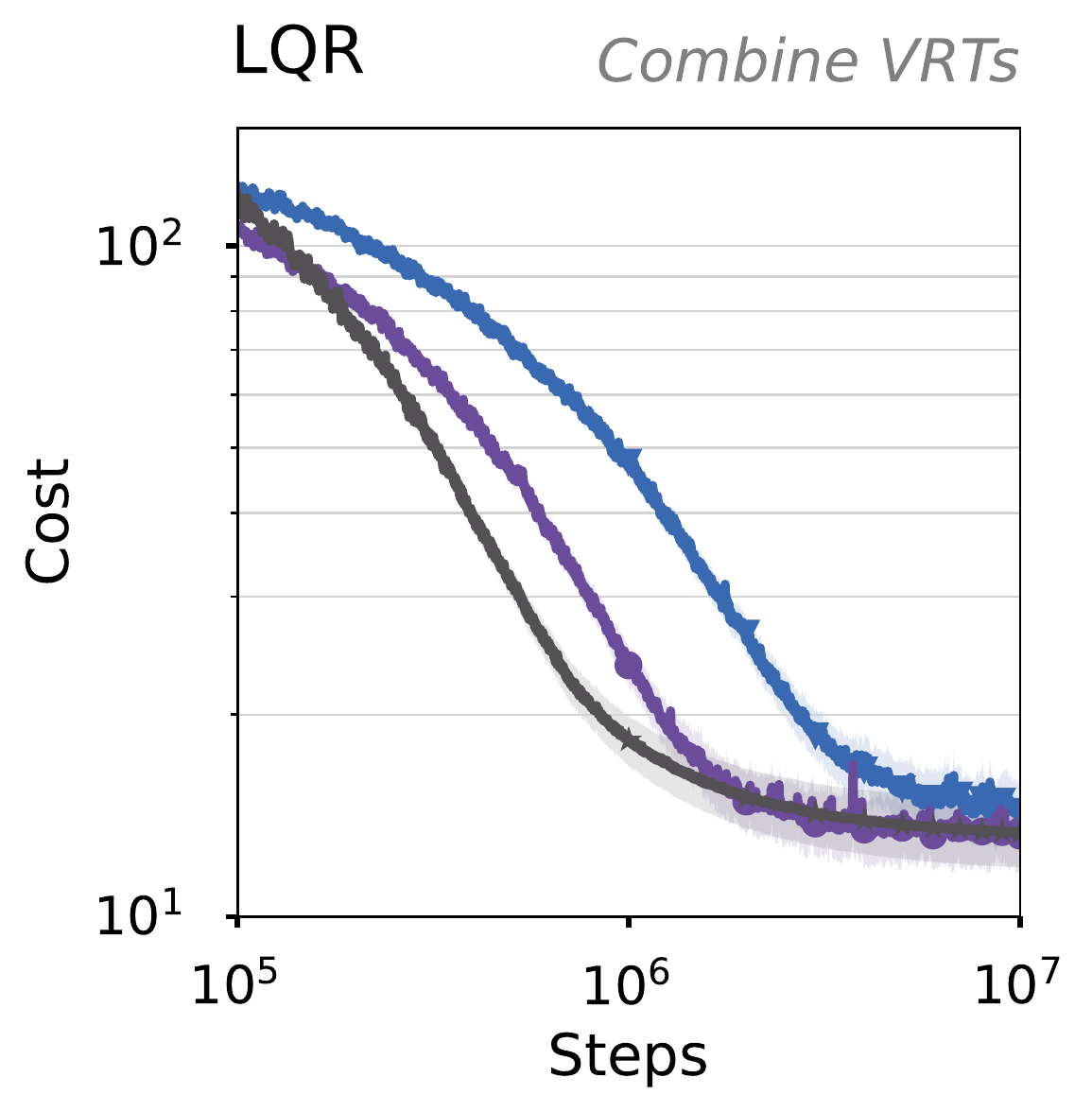} \\
        \includegraphics[width=0.59\linewidth]{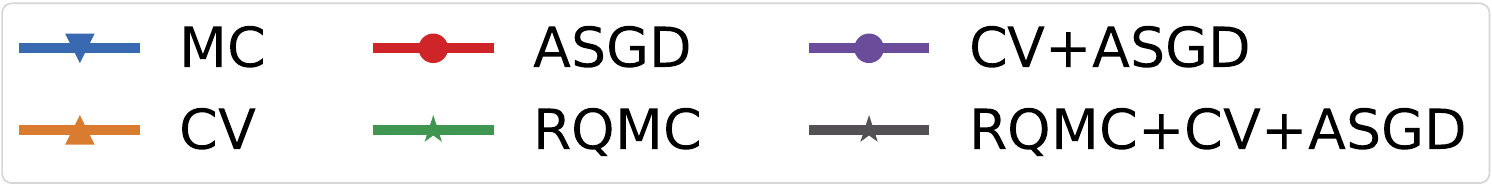}
    \end{center}
    \caption{\small 
        \textbf{RQMC outperforms but also complements other variance reduction techniques.}
        Left: On the LQR, RQMC matches or outperforms control variates (CV) and Accelerated SGD (ASGD;~\citet{jain2018accelerating}).
        Right: Combining those variance reduction techniques performs best, showing their complementarity.
    }
    \label{fig:exp-vrts-main}
    \vspace{-1.0em}
\end{figure}

%% file: related.tex
\section{RELATED WORK}\label{sec:related}

\paragraph{Randomized Quasi Monte-Carlo Methods}
Randomized Quasi-Monte Carlo (RQMC) methods are used throughout science and engineering to improve estimation of intractable integrals.
Previous application domains of RQMC include finance~\citep{Glasserman2013-zq,joy1996quasi, lecuyer2005quasi}, computational biology~\citep{Beentjes2019-mj, Puchhammer2021-og}, and statistics and machine learning~\citep{Gerber2015-lb, pmlr-v80-buchholz18a, Liu2021-gu}.
From a theoretical perspective, \citet{LEcuyer2008-tm} show that RQMC can reduce the variance in Markov chain settings and they propose \emph{Array-RQMC}, an RQMC method taylored to Markov chains with long horizons (see the \supp for preliminary experiments).
Up to our knowledge, this work is the first that investigates RQMC in the context of reinforcement learning.
For a more thorough review on the details of RQMC, we refer to~\citep{Niederreiter_1992-gy, LEcuyer2016-hj}.

\paragraph{Policy Gradient and Continuous Control}
Policy gradient methods~\citep{williams1992simple, sutton2000policy} are popular for continuous control problems as they naturally handle continuous action-space problems.
A early applications of policy gradients to continuous control type problems (including LQR) can be found in \citet{benbrahim1997biped} and \citet{kimura1998reinforcement}.
Subsequently, notable applications include the control of a humanoid biped~\citep{peters2012policy}, dexterous manipulation~\citep{Rajeswaran-RSS-18}, quadrupedal locomotion~\citep{kohl2004policy}, and simulated car driving from high-dimensional inputs~\citep{wierstra2007solving}.
\citet{peters2008reinforcement} provide a review.

Specific to variance reduction, \citet{Weaver2001-qm} and~\citet{Greensmith2004-gc} derive bounds for optimal control variates (CV) on immediate and expected returns.
Those optimal baselines are often hard to estimate, and more generally replaced with (approximate) state-value functions~\citep{mnih2016asynchronous, schulman2015high}.
Several works have proposed to further condition the CV on the actions~\citep{Grathwohl2017-mj, liu2018actiondependent}, but it is unclear if this those kinds of baselines are beneficial~\citep{tucker2018mirage}.
\citet{Romoff2018-dd} show that learning a reward function in addition to a baseline effectively reduces gradient variance for environment with noisy rewards.
From an optimization perspective, \citet{papini2018stochastic} adopt a variance-reduced optimizer (SVRG;~\citep{Johnson2013-lb}) for policy improvement.
\citet{Du2017-oo} propose to also use SVRG to learn a state-value function, effectively tackling a policy evaluation problem.
\citet{Peng2019-rn} further extended this approach to the mini-batch setting.

RQMC is orthogonal to all the above approaches and can be combined with them, see \S\ref{sec:experiments}. Additionally, it has several additional appealing properties.
First, the same approach (and code) equally handles policy evaluation and learning, unlike CV or optimization methods which require different design choices for both.
Second, RQMC is simple to implement and does not require additional hyper-parameters. 
Finally, RQMC retains all appealing factors of policy gradient, remaining universally applicable.

%% file: conclusion.tex
\section{CONCLUSION}\label{sec:conclusion}

We propose to replace the MC steps in policy evaluation and learning with Randomized Quasi-Monte Carlo sampling.
This drop-in sampling technique is compatible with several existing state-of-the-art algorithms and has resulted in improved empirical results in  continuous control problems.
In particular, we observe reduced variances in policy value estimation as well as improved estimation of policy gradients, while reducing the number of samples required to estimate those quantities.

We foresee a few directions for future research.
While our work is empirical in nature, a more formal characterization is needed to understand when RQMC is guaranteed to improve policy learning and evaluation.
For example, it is well-known that RQMC can provably underperform MC on some (contrived) cases where smoothness requirements are not satisfied~\citep{Sloan1998-av}.
Do those cases also arise in reinforcement learning?
Then, from a practical standpoint, our RQMC experiments took approximately 1.2x longer to run than MC ones --- can we address this slow-down with specialized software?
Finally, what role do specialized Markov chain methods (\eg, Array-RQMC~\citep{LEcuyer2008-tm}) play in reinforcement learning?
We hope the promising results presented in this paper can help motivate those lines of inquiry.

%% file: supp/supp_aistats.tex
\onecolumn

\include{supp/experimental_details}
\vfill
\pagebreak

\include{supp/full_experiments}
\vfill
\pagebreak

%% file: supp/experimental_details.tex
\section{EXPERIMENTAL DETAILS}

This section provides additional details on the tasks, algorithms, and implementations in our experimental setups.

Our code (and other resources) is available at: \url{http://seba1511.net/projects/qrl/}

\subsection{MDP BACKGROUND}

All the tasks considered in our experimental section can be modelled as a Markov decision process (MDP).
A MDP can be formalized as a 4-tuple $(\mathcal{S}, \mathcal{A}, T, R)$, where $\mathcal{S}$ and $\mathcal{A}$ are the spaces of states and actions, respectively.
The transition distribution $T(s' \mid s, a)$ indicates the probability of transitioning from state $s \in \mathcal{S}$ to next state $s' \in \mathcal{S}$ when taking action $a \in \mathcal{A}$.
In case the transition dynamics are deterministic (as in some \mujoco tasks), $T$ degenerates to a Kronecker delta function.
Finally, the reward function $R(s, a)$ assigns a reward scalar $r \in \mathbb{R}$ to each state-action pair $(s, a)$, and defines the objective the agent is tasked to optimize.
As we consider finite horizon tasks, we omit possible discount factors and assume they are implicitly absorbed into the reward function.

\subsection{BROWNIAN MOTION}

For the Brownian motion experiments, the agent is a point-mass with position $s \in \mathbb{R}$ and takes actions $a \in \mathbb{R}$.
Given current state $s$ and action $a$, the next state is deterministically computed with $s^\prime = s + 0.1 \cdot a$ where $a \sim \mathcal{N}(\mu, \sigma)$ for state-independent parameters $\mu \in \mathbb{R}$ and $\sigma \in \mathbb{R}$, and the immediate reward is $R(s, a) = \norm{s^\prime}_2$.
The initial state is always $s_1 = 0$ and the episode terminates after $T = 20$ timesteps.

\subsection{LINEAR-QUADRATIC REGULATOR}

In LQR experiments, we use a Gaussian policy $\pi_K(a_t \mid s_t) = \mathcal{N}(Ks_t, I_{6})$ where $K \in \mathbb{R}^{6 \times 8}$ is a learnable matrix and $I_6$ is the identity matrix in $\mathbb{R}^6$.
Given initial state $s_1 = \frac{\epsilon_0}{\norm{\epsilon_0}_2}$ with $\epsilon_0 \sim \mathcal{N}(0, I_8)$, states and immediate rewards at timestep $t \in [1, \dots, 20]$ are computed with
\begin{align}
    s_{t+1} &= As_t + Ba_t + \epsilon_t \qquad \text{ where } \quad \epsilon_t \sim \mathcal{N}(0, \Sigma_s) \\
    R(s_t, a_t) &= - s_t^\top P s_t - a_t^\top Q a_t,
\end{align}
where $A \in \mathbb{R}^{8 \times 8}$ and $B \in \mathbb{R}^{8 \times 6}$ are constructed by first sampling entries from a standard normal distribution $\mathcal{N}(0, 1)$ and then normalizing the matrices to have unit Frobenius norm.
Matrices $P \in \mathbb{R}^{8 \times 8}$ and $Q \in \mathbb{R}^{6 \times 6}$ are random positive semi-definite and constructed to have unit condition number.
The state covariance matrix $\Sigma_s$ is also random positive semi-definite with unit condition number.

\subsection{MUJOCO}

We use the standardized \mujoco tasks (\swimmer, \cheetah, \hopper, \walker, and \ant) as described by~\citet{brockman2016openai}.
Specifically, we use the implementations in \texttt{gym} version \texttt{0.20.0}, \texttt{mujoco-py} version \texttt{1.50.1.68}, and \texttt{\mujoco} version \texttt{1.50}.

\subsection{METHODS DETAILS}

This section provides more details on learning algorithms.
As the VPG implementation for Brownian motion and LQR tasks closely follows the presentation in the main text, we focus on our implementation of SAC.

The two main components of SAC are the loss functions for $\hQ$ and $\policy$ augmented with a maximum entropy objective.
To learn $\hQ$, SAC minimizes the on-policy squared Bellman error
\begin{equation}
    \mathcal{L}_\hQ = \Exp[s_t, a_t, s_{t+1}]{\left(\bot\left(\Exp[a_{t+1}]{\hQ(s_{t+1}, a_{t+1}) - \alpha \log \policy(a_{t+1} \vert s_{t+1})} + R(s_t, a_t)\right) - \hQ(s_t, a_t) \right)^2},
\end{equation}
where $\bot(\cdot)$ is the stop-gradient operator, and $\alpha \in \mathbb{R}$ is the weight of the entropy bonus encouraging exploration.
In the above equation, $s_t, a_t, s_{t+1}$ are sampled from a replay buffer of past experience while $a_{t+1}$ is freshly sampled for each evaluation of $\mathcal{L}_\hQ$.
While most implementations use a single action to estimate the inner expectation $\mathbb{E}_{a_{t+1}}$ over next actions $a^\prime$, we found 8 actions to work as well or better with MC and performed significantly better with RQMC.

To optimize the policy $\policy$, SAC maximizes the expected returns as approximated by $\hQ$ (\cref{eq:qmcac-value}).
The policy loss augmented with a max-entropy term is given by
\begin{equation}
    \mathcal{L}_\policy = - \Exp[s]{\Exp[a]{\hQ(s, a) - \alpha \log \policy(a \mid s)}},
\end{equation}
where $s$ is sampled from a replay buffer, and $a$ from the current policy $\policy$.
In this case too, we found it beneficial to use 8 actions for the expectation over actions $\mathbb{E}_{a}$ for a given state $s$.

For more details, please refer to the provided code implementation.

\subsection{LEARNING HYPER-PARAMETERS}

\begin{table}[t]
\vspace{-1em}
\centering
\makebox[0pt][c]{\parbox{1.2\textwidth}{%
    \centering
    \hfill
    \begin{minipage}[b]{0.42\hsize}\centering

        \renewcommand{\arrayrulewidth}{1.3}
        \centering
        \caption{\small
            Hyper-parameters for LQR tasks.
            \label{tab:lqr-hp}
        }
        \setlength{\tabcolsep}{3pt}
        \vspace{-1em}
        {\small
        \begin{tabular}{@{}l r@{}}
            \addlinespace
            \toprule
                Hyper-parameter & Value \\
            \midrule

                Learning Rate                              & $0.0007$ \\
                Momentum                                   & $0.99$ \\
                Trajectories / Updates                     & $16$ \\
                CV's $\gamma$ (discount)                   & $0.99$ \\
                CV's $\lambda$ (GAE interpolation)         & $0.95$ \\
                ASGD's $\kappa$ (long to short step ratio) & $1000.0$ \\
                ASGD's $\xi$ (statistical advantage)       & $10.0$ \\

            \bottomrule
        \end{tabular}
        }

    \end{minipage}
    $\qquad$
    \begin{minipage}[b]{0.42\hsize}\centering

        \vspace{-1em}
        \renewcommand{\arrayrulewidth}{1.3}
        \centering
        \caption{\small
            Hyper-parameters for \mujoco tasks.
            \label{tab:sac-hp}
        }
        \setlength{\tabcolsep}{3pt}
        \vspace{-1em}
        {\small
        \begin{tabular}{@{}l r@{}}
            \addlinespace
            \toprule
                Hyper-parameter & Value \\
            \midrule

                Learning Rate             & $0.001$ \\
                Entropy Regularization    & $0.2$ \\
                Mini-batch Size           & $100$ \\
                Minimum Replay Size       & $4000$ \\
                Maximum Replay Size       & $10^6$ \\
                MLP Depth ($\policy$ \& $\hQ$)           & $2$ \\
                MLP Width ($\policy$ \& $\hQ$)           & $256$ \\

            \bottomrule
        \end{tabular}
        }

    \end{minipage}
    \hfill
    \hfill
}}
\vspace{-1em}
\end{table}

\paragraph{LQR}
We report hyper-parameters for the LQR experiments (including those combining other variance reduction techniques) in \cref{tab:lqr-hp}.

\paragraph{\mujoco}
We use the same hyper-parameters for all tasks and algorithms (SAC, TD3, and DDPG) on \mujoco tasks, given in \cref{tab:sac-hp}.

\subsection{CODE SNIPPETS}

This section describes how to implement RQMC with popular software packages.
For the Brownian motion and LQR experiments, we use the left matrix scramble and digital shift implementation in SSJ~\citep{iLEC16j}, and for the \mujoco tasks we use Owen's scrambling as implemented in PyTorch~\citep{paszke2019pytorch}.
Pseudocodes are listed in \cref{alg:ssj-lms} and \cref{alg:pytorch-owen}.
A complete implementation of SAC with RQMC (built on \texttt{Spinning-Up}~\citep{SpinningUp2018}) is included with the supplementary material.

\begin{algorithm}
    \caption{Left Matrix Scramble and Digital Shift in SSJ~\citep{iLEC16j}}
    \label{alg:ssj-lms}
    \vspace{-0.5em}
    \begin{lstlisting}[escapeinside={(*}{*)}, language=Java]
public double[][] sample(int pow, int dim) {
    MRG32k3a stream = new MRG32k3a();
    int n_samples = (int)Math.pow(2, pow);
    double[][] pointset = new double[n_samples][dim];

    DigitalNetBase2 p = new SobolSequence(pow, 31, dim);
    p.leftMatrixScramble(stream);
    p.addRandomShift(stream);

    PointSetIterator point_stream = p.iterator ();
    for (int i = 0; i < n_samples; ++i) {
        point_stream.nextPoint(pointset[i], dim);
    }
    return pointset; // Use pointset as in l. 4 - 10 of (*{\color{gray}\cref{alg:rqmc_pg}}*)
}

    \end{lstlisting}
    \vspace{-0.5em}
\end{algorithm}

\begin{algorithm}
    \caption{RQMC Policy Evaluation w/ Critic in PyTorch~\citep{paszke2019pytorch}}
    \label{alg:pytorch-owen}
    \vspace{-0.5em}
    \begin{lstlisting}[escapeinside={(*}{*)}, language=Python]
from torch.quasirandom import SobolEngine
rqmc = SobolEngine((*$\dimA$*), scrambled=True)
(*$R$*) = 0.0
for (*$k$*) in (*$1, \dots, M$*):  # evaluation on M states
    (*$s_k$*) = replay_buffer.states[(*$k$*)]
    rqmc._scramble()
    rqmc.reset()
    (*$u^{(k)}$*) = rqmc.draw((*$N$*))  # (*{\color{gray}$u^{(k)} \in \mathbb{R}^{N \times \dimA}$}*)
    for (*$j$*) in (*$1, \dots, N$*):  # N actions per state
       (*$a^{(k)}_j$*) = (*$\mu_\params(s_k) + \sigma_\params(s_k) \odot F^{-1}(u^{(k)}_j)$*)
       (*$R$*) += (*$\Q(s_k, a^{(k)}_j)$*)
return (*$R\, /\, (M \cdot N)$*)
    \end{lstlisting}
    \vspace{-0.5em}
\end{algorithm}

%% file: supp/full_experiments.tex
\section{ADDITIONAL EXPERIMENTAL RESULTS}

This section presents experiments that supplement the ones from the main text.
In particular, it includes results for all \mujoco tasks (\swimmer, \cheetah, \hopper, \walker, \ant), and an extension of RQMC (\arqmc) specifically designed for Markov chains.

\subsection{POLICY EVALUATION}

\begin{figure}[H]
    \begin{center}
        \includegraphics[width=0.32\linewidth]{figs/policy_evaluation/H20_brownian.pdf}
        \includegraphics[width=0.32\linewidth]{figs/policy_evaluation/H20_lqr.pdf}
        \includegraphics[width=0.32\linewidth]{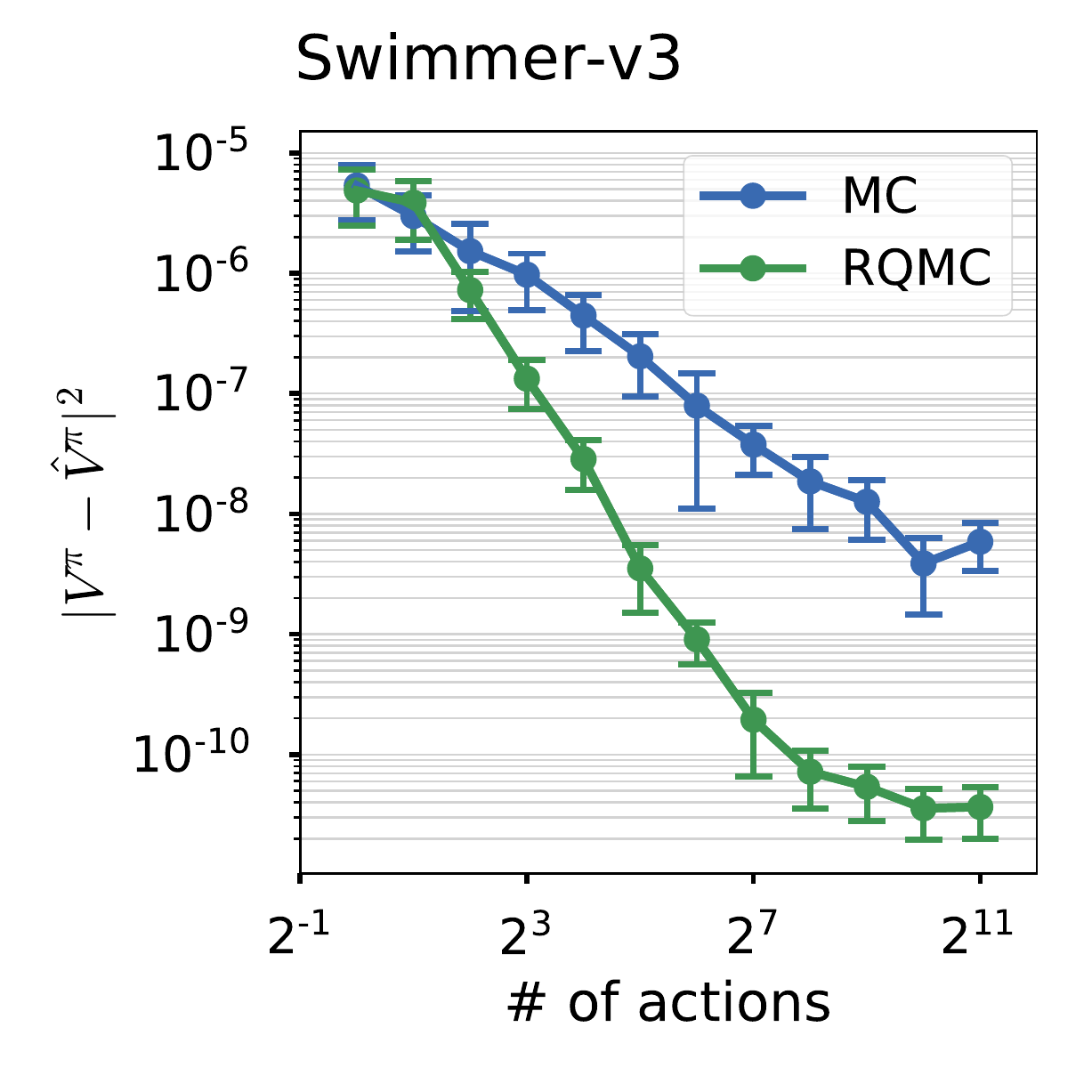}
        \includegraphics[width=0.32\linewidth]{figs/policy_evaluation/sac_cheetah.pdf}
        \includegraphics[width=0.32\linewidth]{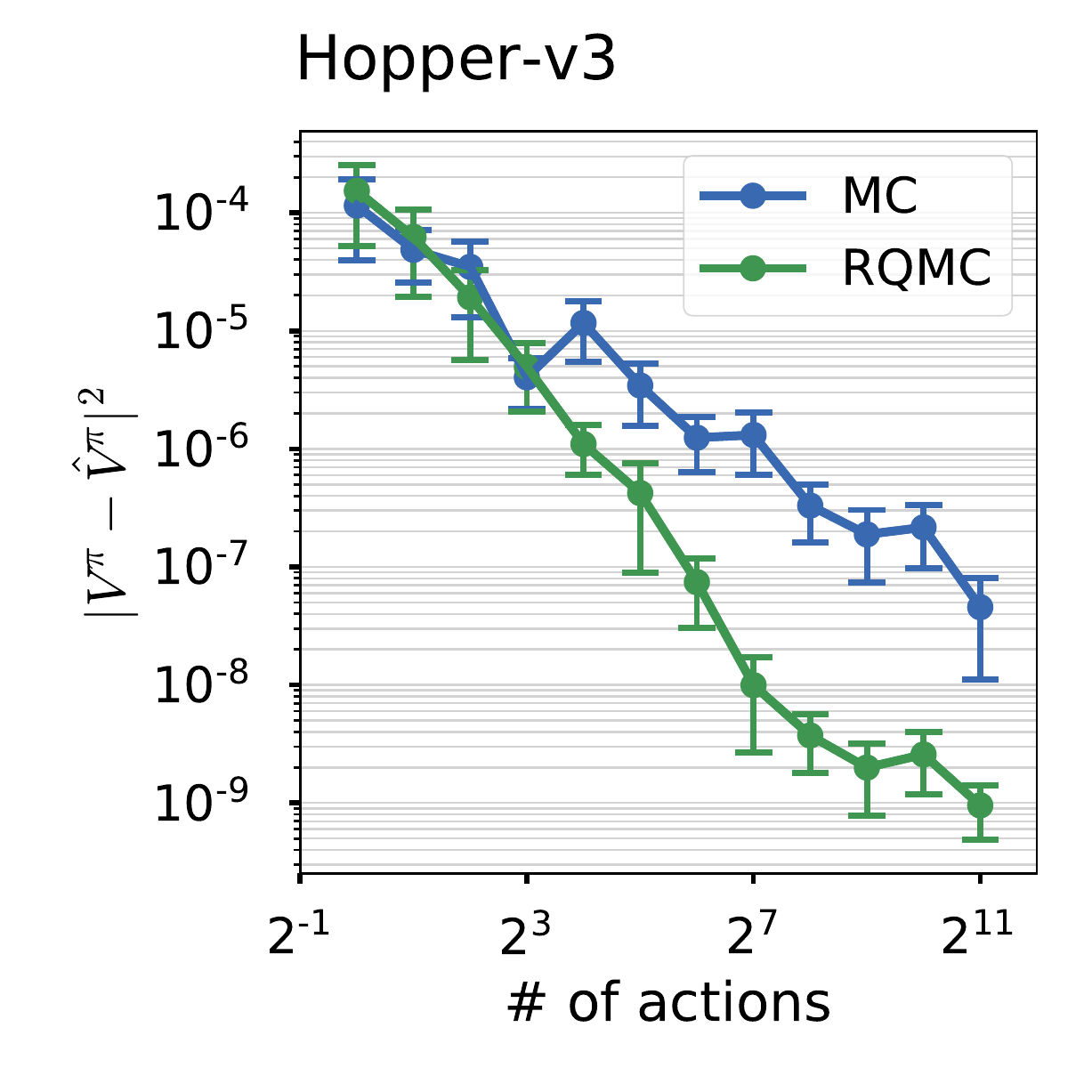}
        \includegraphics[width=0.32\linewidth]{figs/policy_evaluation/sac_walker.pdf}
        \includegraphics[width=0.32\linewidth]{figs/policy_evaluation/sac_ant.pdf}
    \end{center}
    \caption{\small 
        \textbf{RQMC reduces value estimator error.}
        RQMC is  much more efficient than MC in estimating the value of a policy on a given number of trajectories.
        As suggested by theory, the gap between RQMC and MC grows with the number of trajectories.
        For Mujoco tasks where ground-truth gradient is not available, we approximate it using $2^{16}$ actions.
    }
    \label{fig:exp-pe-full}
\end{figure}

\cref{fig:exp-pe-full} reports value estimation error akin to \cref{fig:exp-pe-main} of the main text. 
On the two new \mujoco tasks, we observe again that:
\begin{enumerate}
    \item RQMC significantly reduces value estimation error (by up to 2 orders of magnitude on \swimmer), even with relatively few (16) trajectories or actions (128  on \walker, 16 on others).
    \item RQMC converges more rapidly than MC in value estimation error.
\end{enumerate}

\subsection{POLICY IMPROVEMENT}

\begin{figure}[H]
    \begin{center}
        \includegraphics[width=0.32\linewidth]{figs/policy_learning/returns_reinforce_lqr.pdf}
        \includegraphics[width=0.32\linewidth]{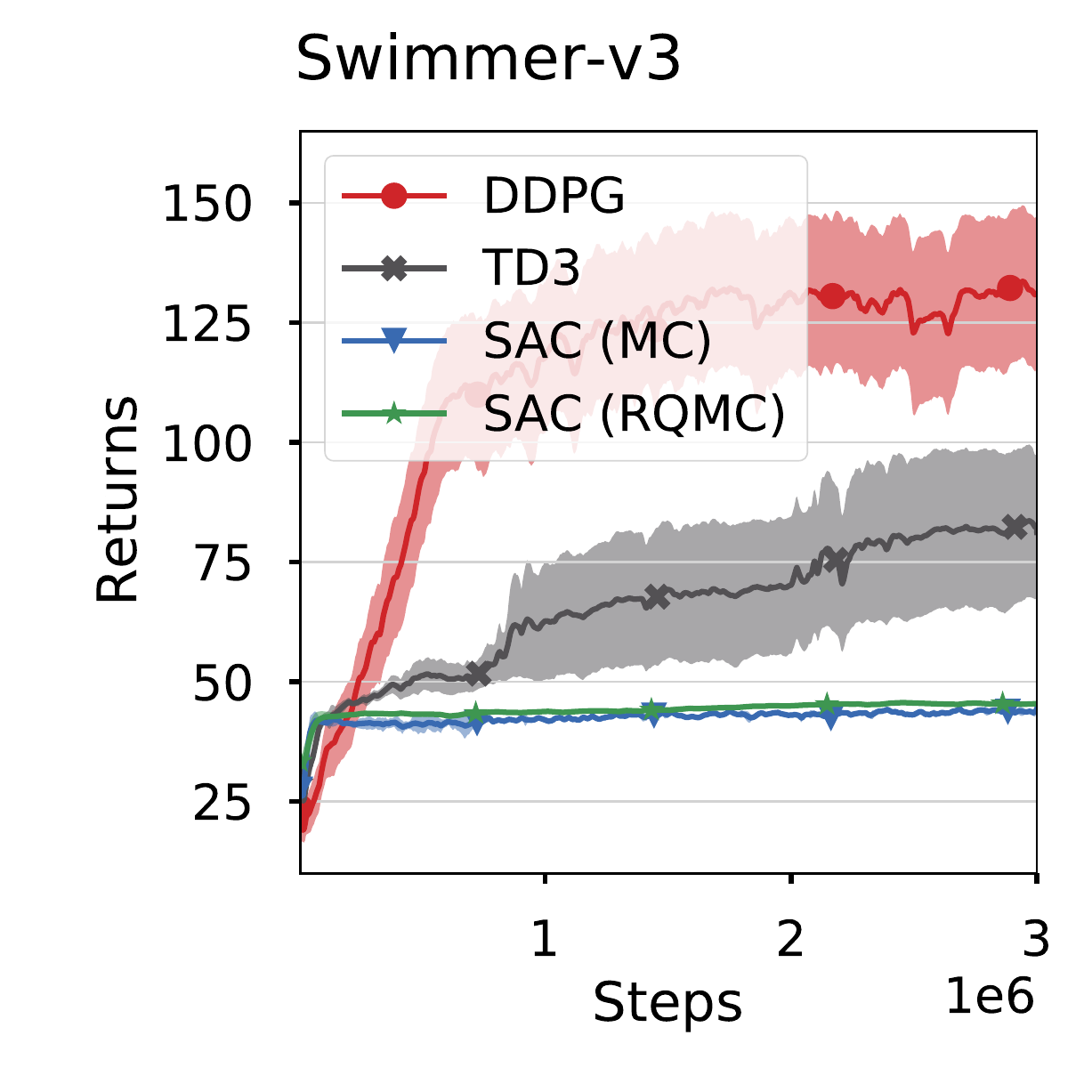}
        \includegraphics[width=0.32\linewidth]{figs/policy_learning/returns_sac_cheetah.pdf}
        \includegraphics[width=0.32\linewidth]{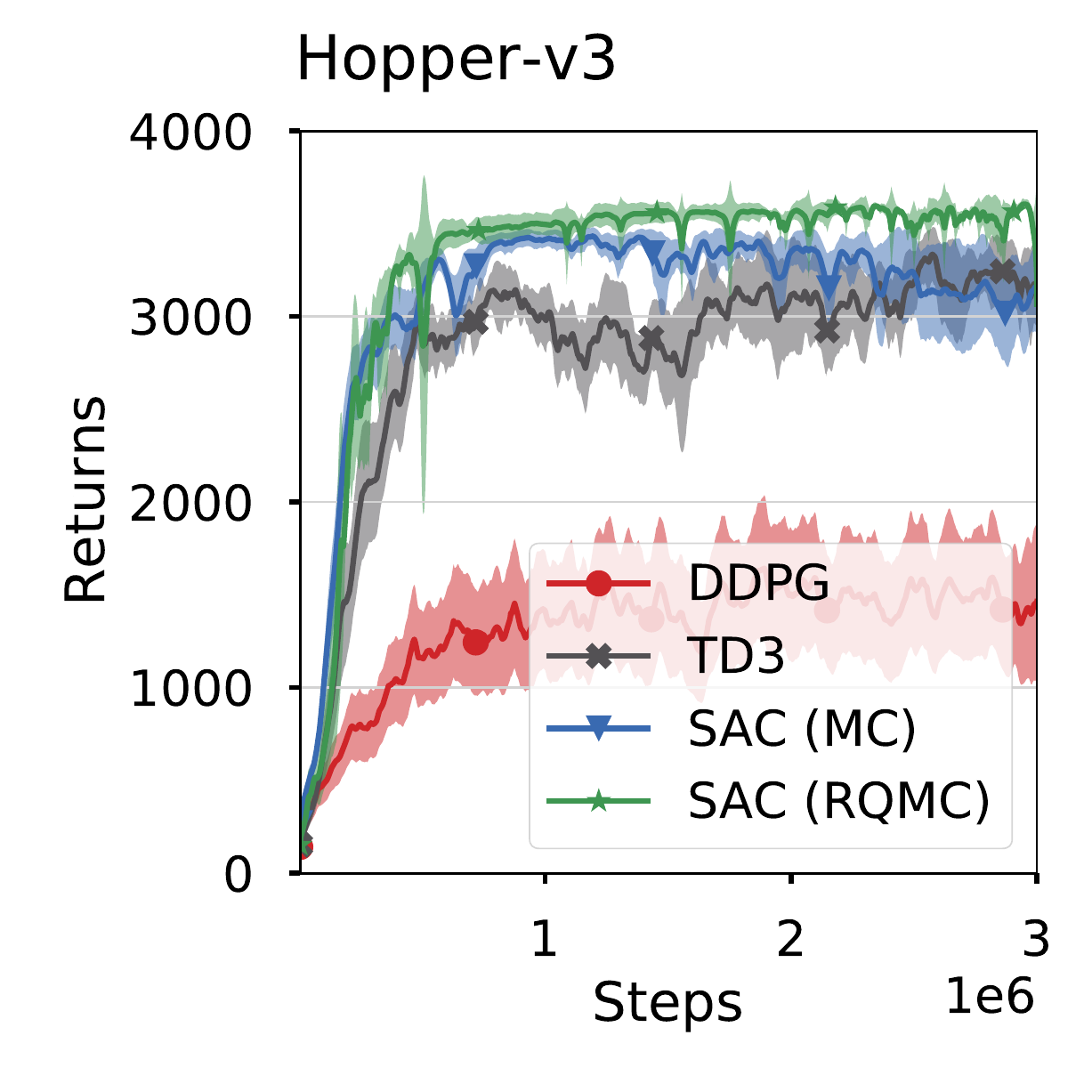}
        \includegraphics[width=0.32\linewidth]{figs/policy_learning/returns_sac_walker.pdf}
        \includegraphics[width=0.32\linewidth]{figs/policy_learning/returns_sac_ant.pdf}
        \caption{\small 
            \textbf{RQMC improves policy learning.}
            Using RQMC during policy learning outperforms MC on LQR and Mujoco tasks.
            In particular, RQMC improves upon MC with SAC -- a state-of-the-art actor-critic method -- in terms asymptotic performance on all Mujoco tasks.
            Unsurprisingly, RQMC alone does not suffice to fix SAC's poor performance on \swimmer where it is surpassed by simpler methods (\ie, DDPG, TD3).
        }
        \label{fig:exp-pi-full}
    \end{center}
\end{figure}

\cref{fig:exp-pi-full} complements \cref{fig:exp-pi-main} from the main text, with policy improvement curves for \swimmer and \hopper.
We observe:
\begin{enumerate}
    \item
        RQMC improves the performance of SAC compared to MC on both \swimmer and \hopper.
        Typically, this improvement is on the same order as upgrading the learning algorithm from TD3 to SAC.
    \item
        RQMC is no panacea: when SAC (MC) fails, adding RQMC does not perform much better.
        This is seen on \swimmer, where simpler methods (DDPG and TD3) drastically outperform SAC with MC or RQMC.
\end{enumerate}

\vfill

\subsection{IMPROVED GRADIENT ESTIMATION}

\begin{figure}[h]
    \begin{center}
            \includegraphics[width=0.32\linewidth]{figs/ablation/gradient_variance/grad_var_lqr.pdf}
            \includegraphics[width=0.32\linewidth]{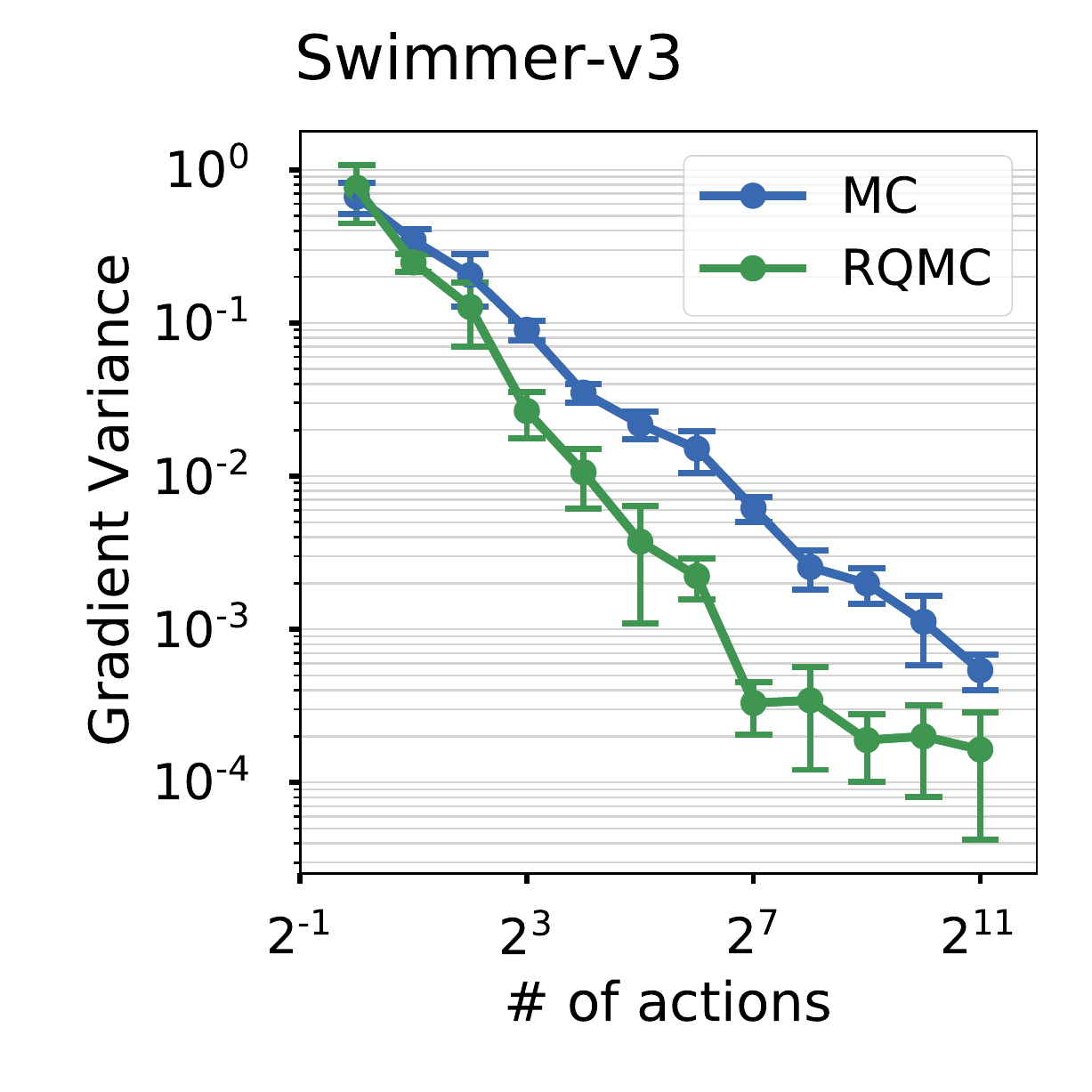}
            \includegraphics[width=0.32\linewidth]{figs/ablation/gradient_variance/grad_var_cheetah.pdf}
            \includegraphics[width=0.32\linewidth]{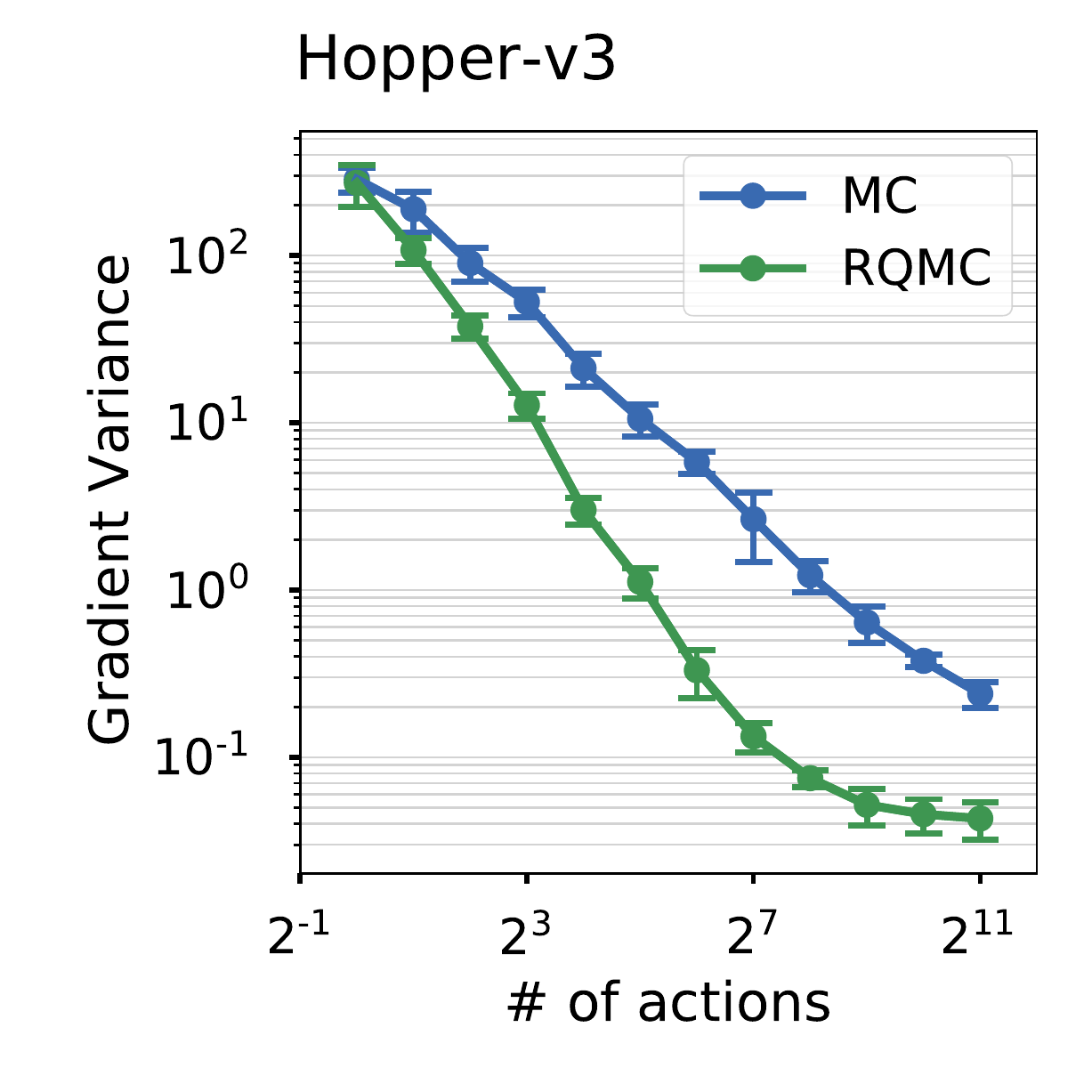}
            \includegraphics[width=0.32\linewidth]{figs/ablation/gradient_variance/grad_var_walker.pdf}
            \includegraphics[width=0.32\linewidth]{figs/ablation/gradient_variance/grad_var_ant.pdf}
    \end{center}
    \caption{\small
        \textbf{RQMC reduces gradient variance.}
        On both LQR and \mujoco tasks, RQMC achieves lower gradient variance than MC for the same number of trajectories.
        Here, variance refers to the trace of the gradient covariance matrix.
        The $95\%$ confidence intervals are computed over 30 random seeds.
    }
    \label{fig:exp-grad-var-full}
    \vspace{-0.2 in}
\end{figure}

\begin{figure}
    \vspace{-1.0em}
    \begin{center}
        \includegraphics[width=0.32\linewidth]{figs/ablation/gradient_alignment/grad_align_lqr.pdf}
        \includegraphics[width=0.32\linewidth]{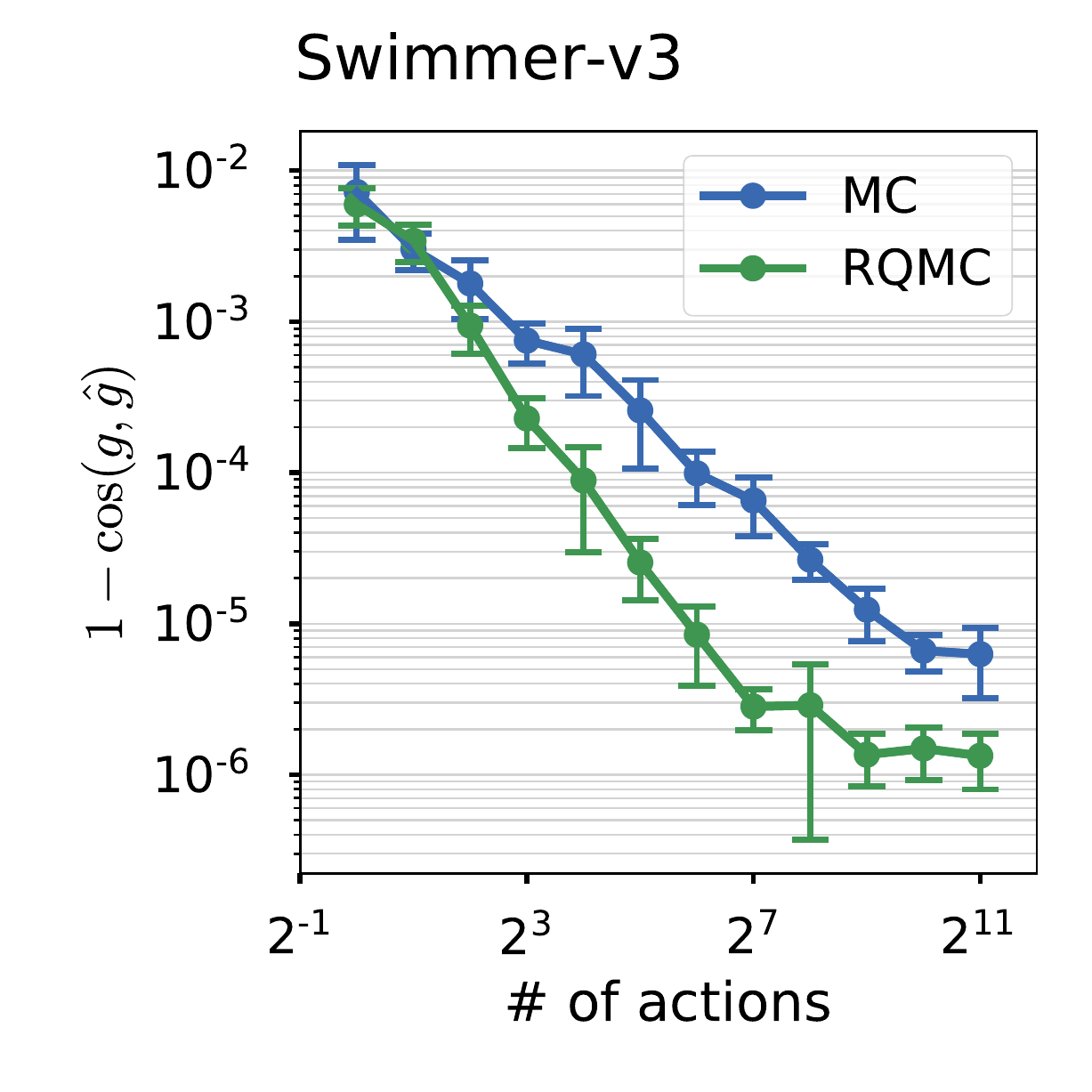}
        \includegraphics[width=0.32\linewidth]{figs/ablation/gradient_alignment/grad_align_cheetah.pdf}
        \includegraphics[width=0.32\linewidth]{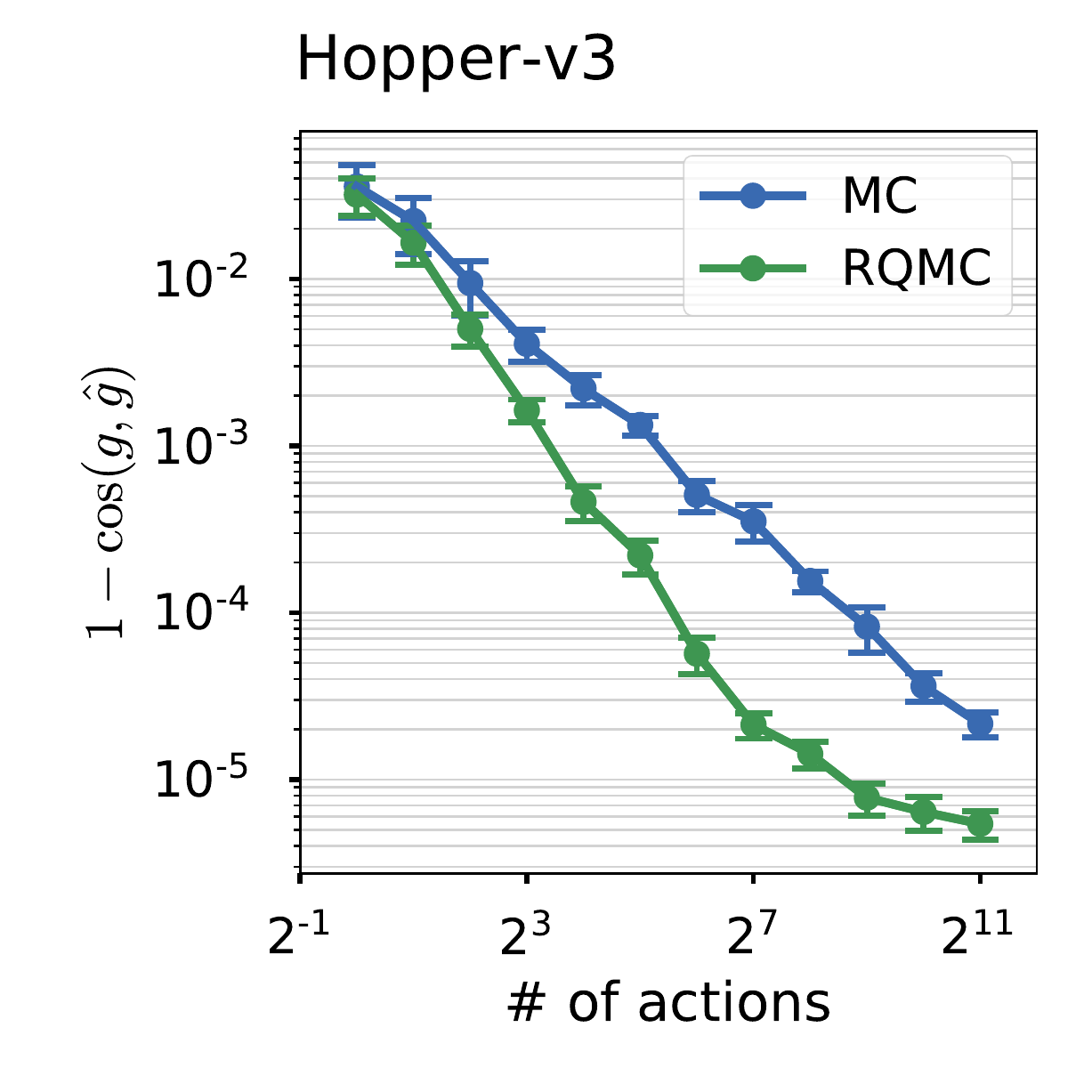}
        \includegraphics[width=0.32\linewidth]{figs/ablation/gradient_alignment/grad_align_walker.pdf}
        \includegraphics[width=0.32\linewidth]{figs/ablation/gradient_alignment/grad_align_ant.pdf}
    \end{center}
    \caption{\small 
        \textbf{RQMC improves gradient alignment.}
        For a given number of trajectories, the gradient direction computed with RQMC is better aligned than when computed with MC.
        The $y$-axis displays the angle between ground-truth and stochastic gradient.
        On LQR, the ground-truth is computed with $48k$ trajectories; on \mujoco, it is estimated using $2^{16}$ actions.
        The $95\%$ confidence intervals are computed over 30 random seeds.
    }
    \label{fig:exp-grad-cos-full}
    \vspace{-1.0em}
\end{figure}

In this subsection, we complete the gradient variance and alignement results of \cref{fig:exp-grad-var-main,fig:exp-grad-cos-main} with \swimmer and \hopper in \cref{fig:exp-grad-var-full,fig:exp-grad-cos-full}.
On both sets of figures, we observe similar trends as in the main text: both gradient variance and alignment improve with more samples (trajectories or actions), and they typically improve at a faster rate with RQMC.

\subsection{ROBUSTNESS TO INCREASED STATE TRANSITION NOISE}

\begin{figure}
    \vspace{-0.0em}
    \begin{center}
        \includegraphics[width=0.24\linewidth]{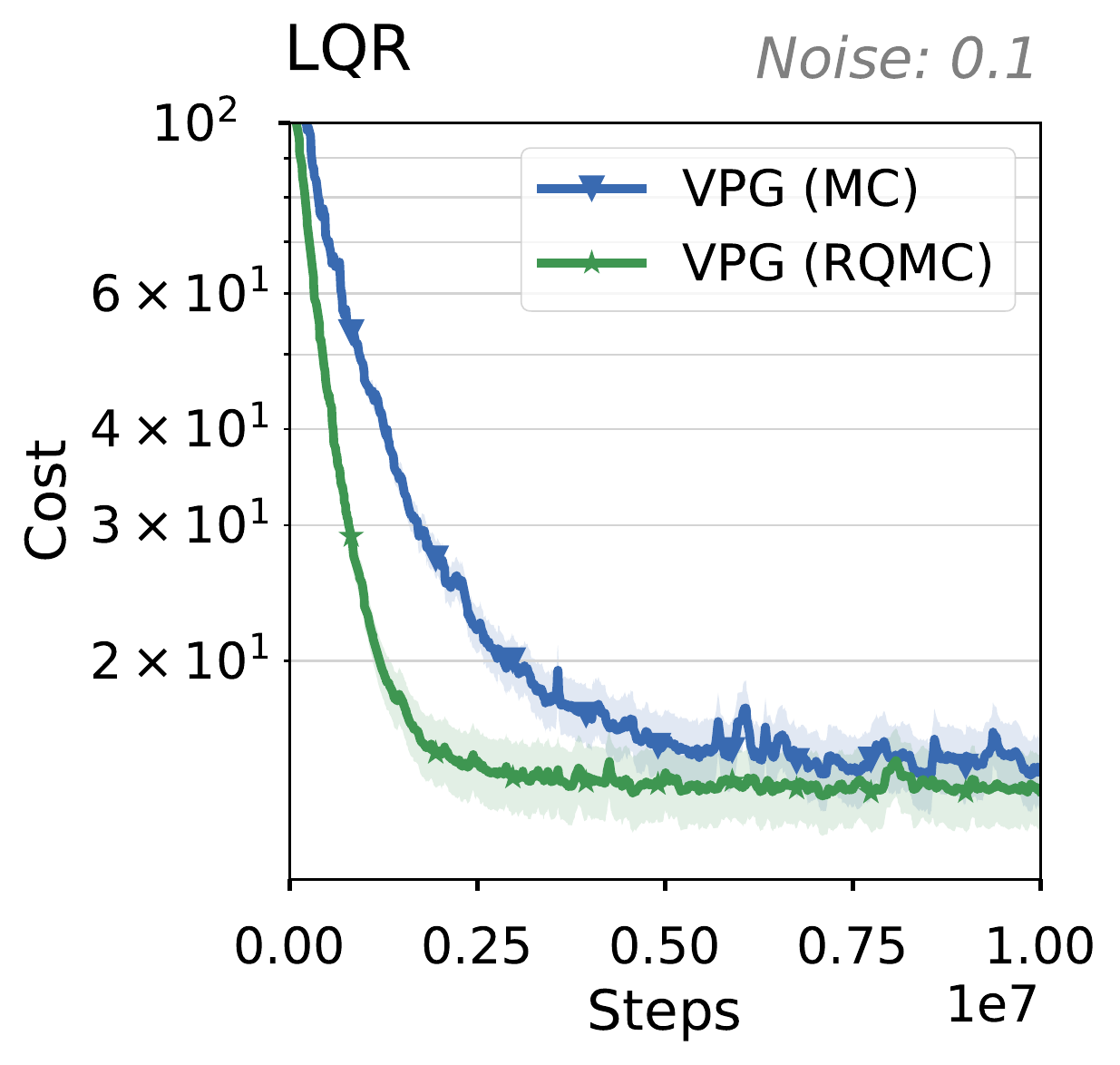}
        \includegraphics[width=0.24\linewidth]{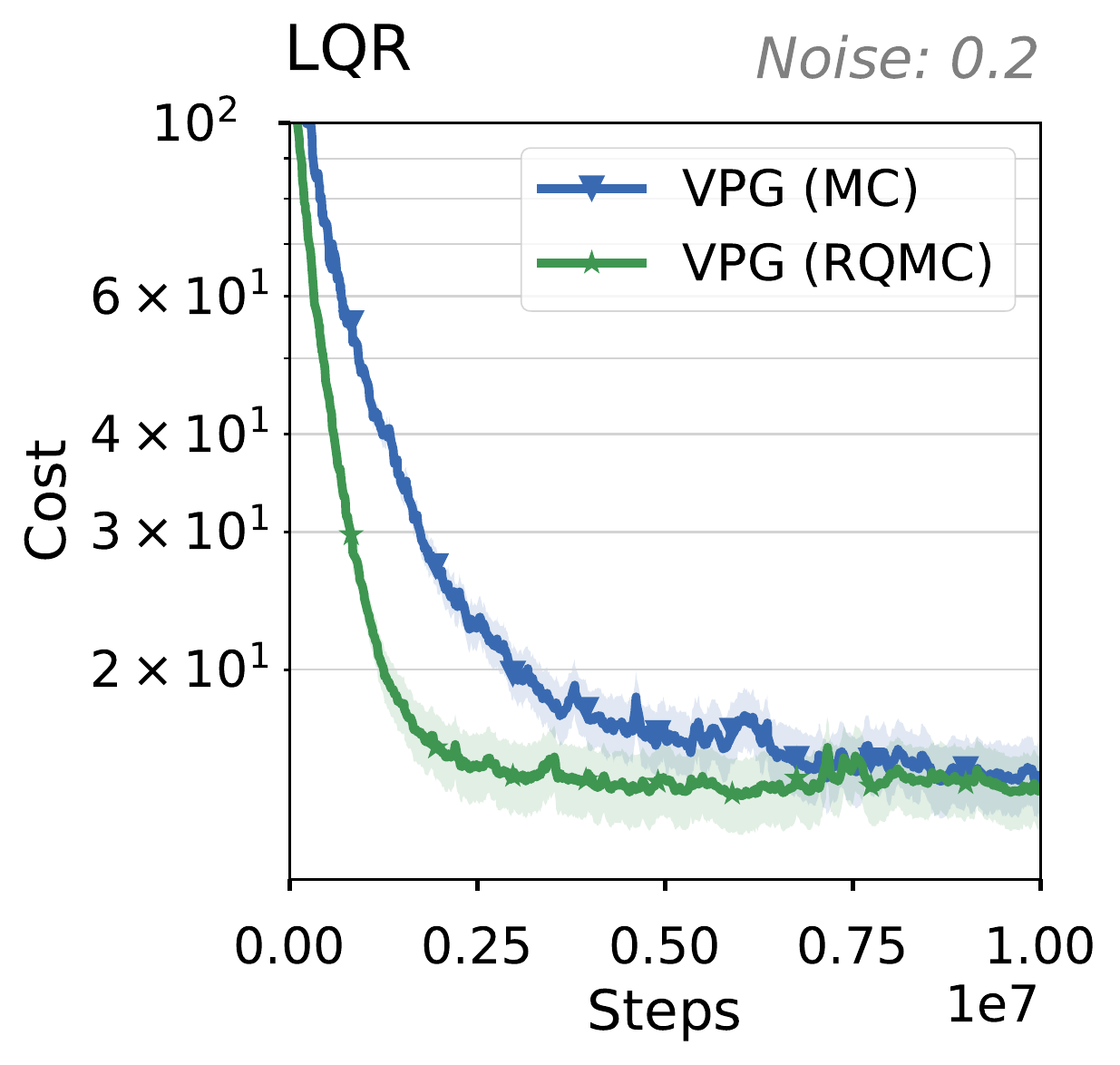}
        \includegraphics[width=0.24\linewidth]{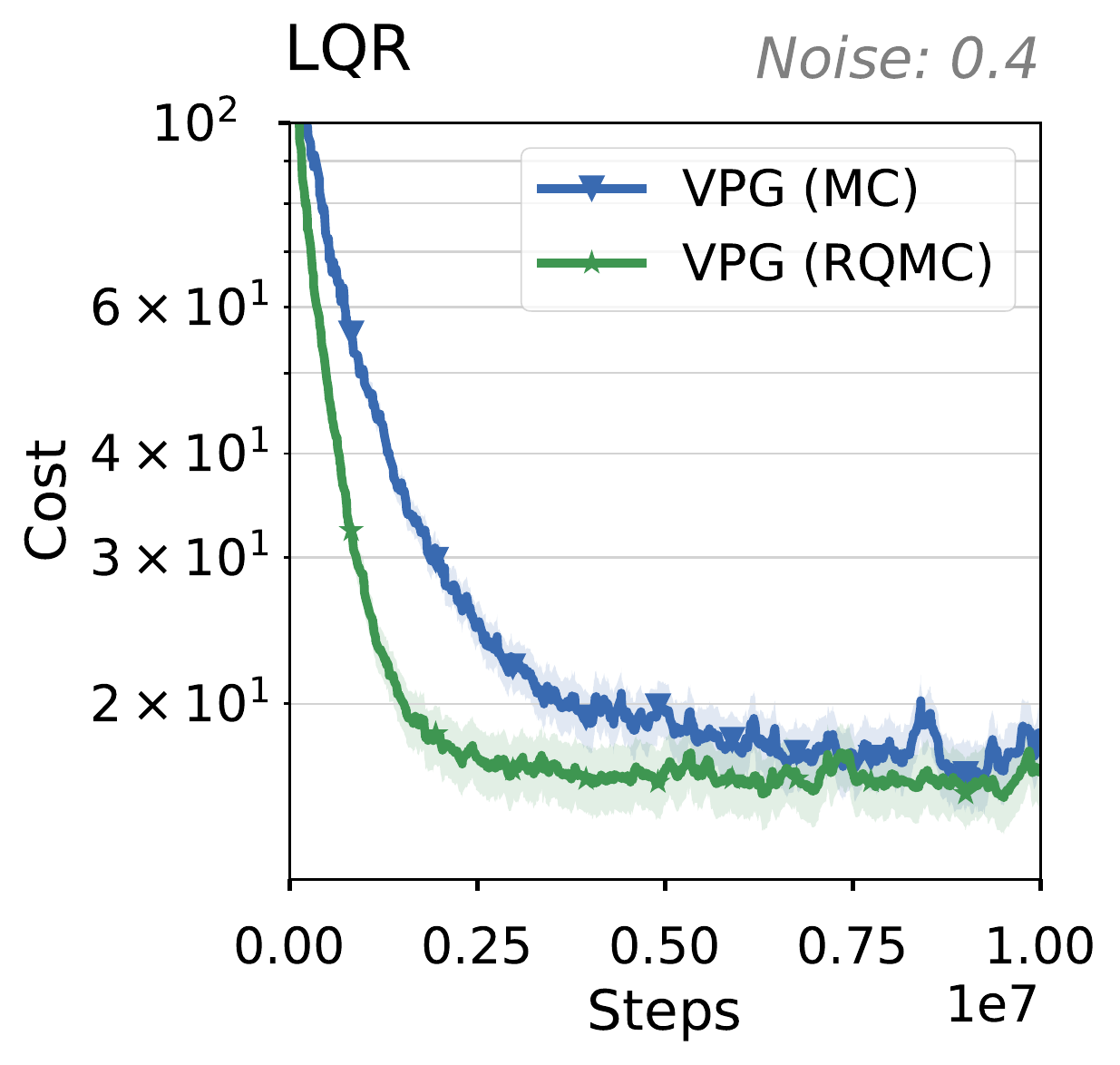}
        \includegraphics[width=0.24\linewidth]{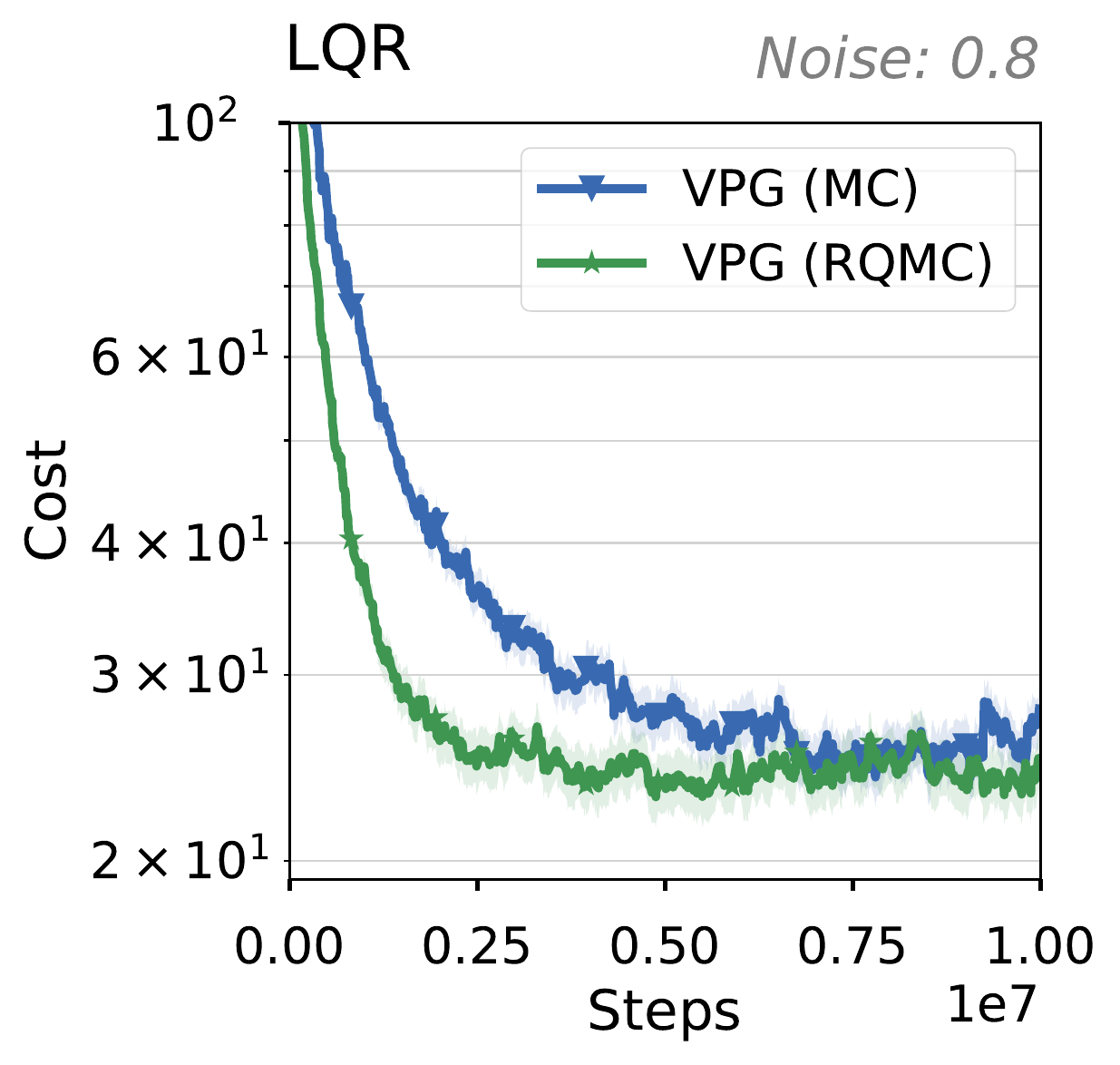}
    \end{center}
    \vspace{-1.0em}
    \caption{
        \label{fig:state-noise}
        \textbf{RQMC outperforms MC under various noise settings.}
        In the LQR setting, we vary the scaling of the state transition covariance noise from $0.1$ to $0.8$.
        Both MC and RQMC perform worse (higher cost) with more noise, but RQMC retains a faster convergence rate than MC on all settings.
    }
    \vspace{-0.0em}
\end{figure}

We now verify the robustness of RQMC under varying state transition noise.
We repeat the VPG experiments on the LQR setting while varying the scaling of the state transition covariance noise from $0.1$ to $0.8$.
\cref{fig:state-noise} reports these results.
As expected, increasing dynamics noise increases asymptotic cost for both MC and RQMC, but RQMC retains its advantage over MC in terms of convergence rate.

\subsection{EXTENSION TO MARKOV CHAINS WITH ARRAY-RQMC}

\begin{figure}
    \vspace{-1.0em}
    \begin{center}
        \includegraphics[width=0.32\linewidth]{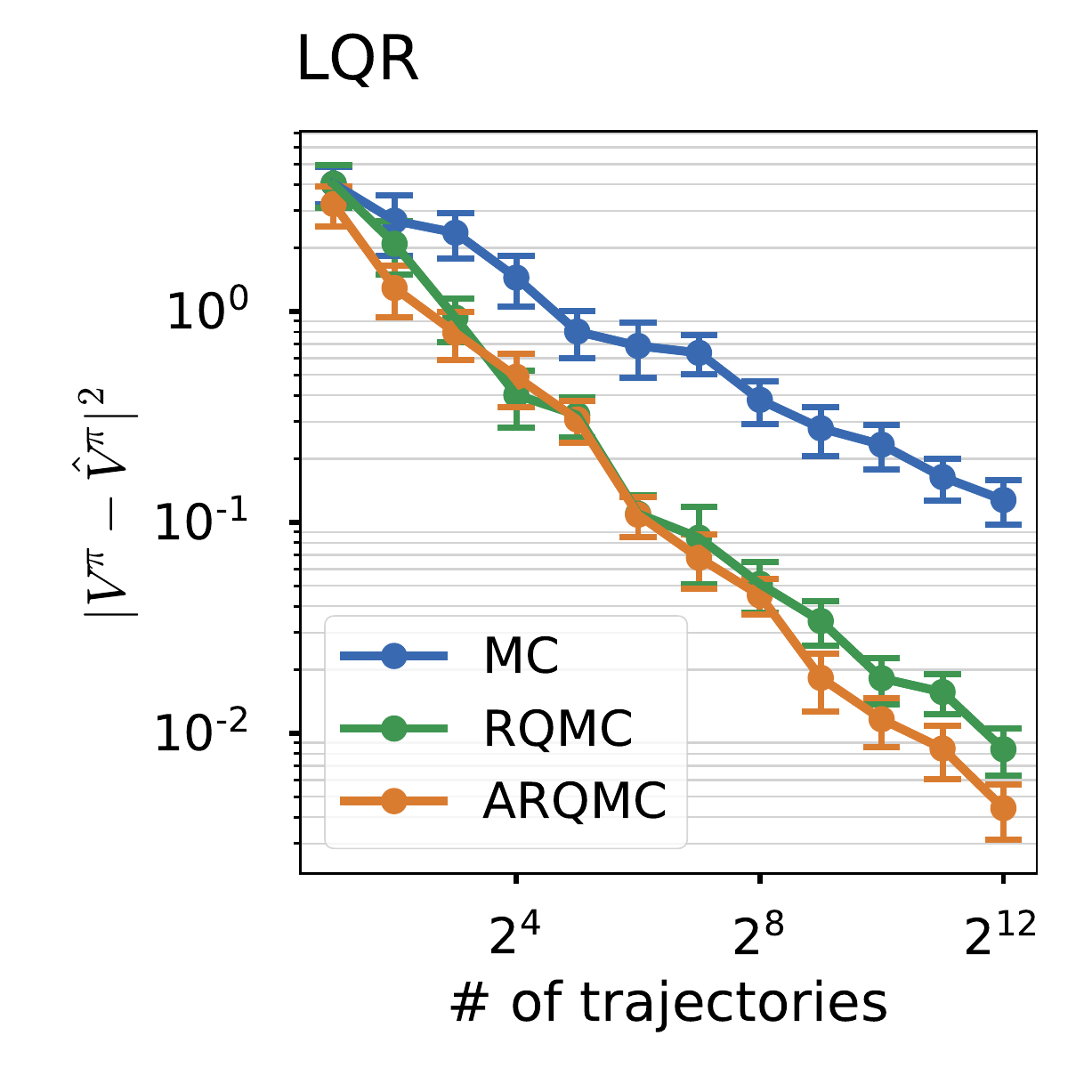}
        \includegraphics[width=0.32\linewidth]{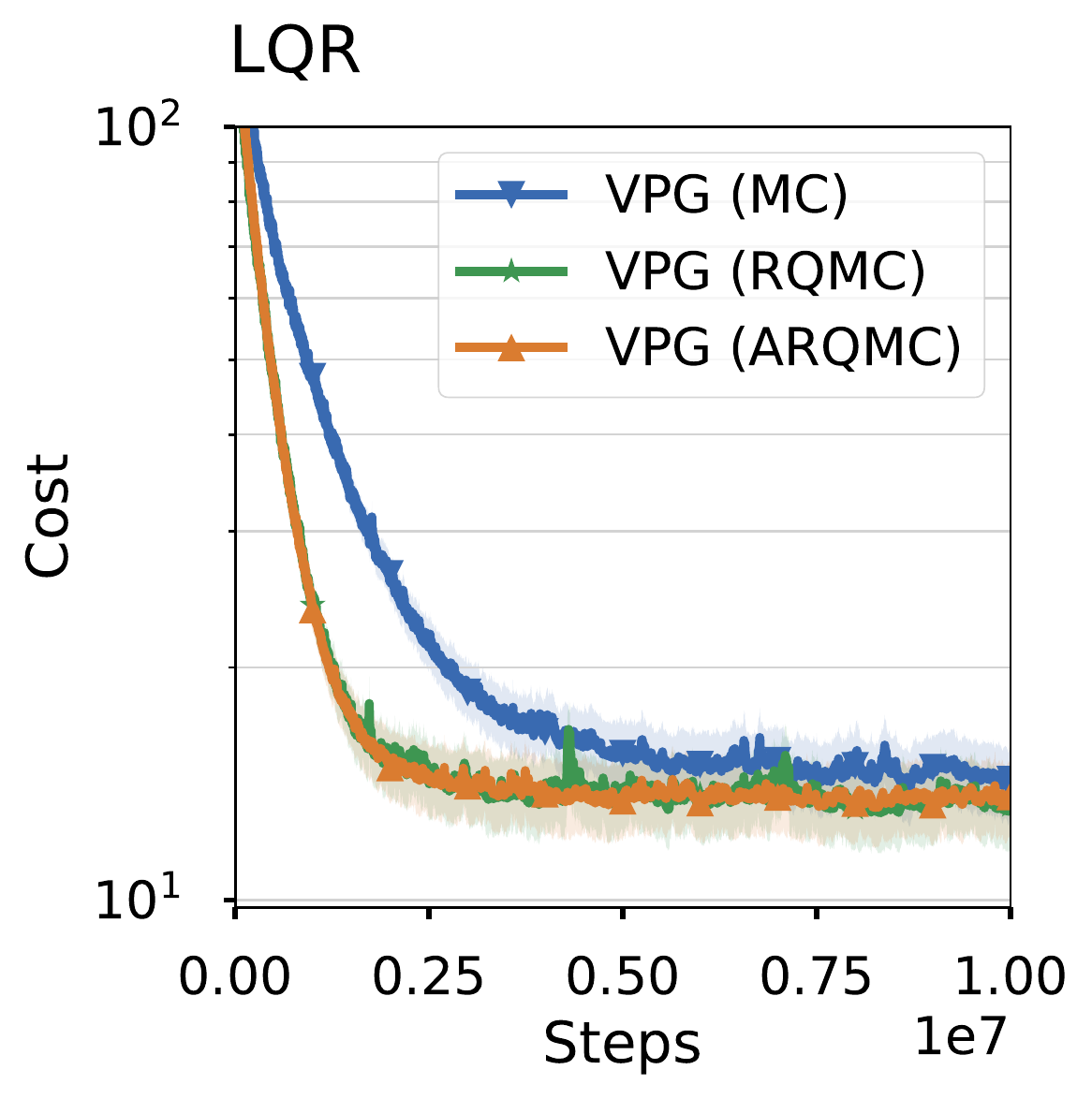}
    \end{center}
    \caption{\small 
        \textbf{Benchmarking RQMC extensions to Markov chains.}
        \arqmc (ARQMC), an RQMC formulation for Markov chains, further improves upon RQMC on the LQR for policy evaluation (left) while closely matching RQMC in learning performance (right).
    }
    \label{fig:exp-arqmc-full}
    \vspace{-1.0em}
\end{figure}

We experiment with \arqmc~\citep{LEcuyer2008-tm, LEcuyer2009-we}, a formulation of RQMC specifically designed for Markov chains.
\arqmc aims to overcome the dependency on the dimension when rolling out a policy for long horizons $T$.
As underlined in the main text, those challenges are twofold: the advantages of RQMC diminish with higher integration dimension, and popular implementations don't support dimensions higher than 21,201.

To address these issues, \arqmc assumes that $M$ trajectories can be collected in parallel, which is almost always the case with simulated environments \citep{pmlr-v119-petrenko20a, brax2021github, Makoviychuk2021isaac}.
Then, \arqmc samples a new RQMC point set $u_1^{(t)}, \dots, u_M^{(t)}$ at every timestep $t$, with one point per trajectory.
Each of the $M$ points $u^{(t)}_m \in \mathbb{R}^{\vert \mathcal{A} \vert}$ has dimensionality $\vert \mathcal{A} \vert$ thus dropping the dependency on $T$.
In order to effectively reduce variance, \arqmc has to carefully assign the current state of each trajectory with points in the RQMC point set.
This assignment is done by sorting all states at timestep $t$ according to an application-dependent value function.
Intuitively, the goal of this sorting function is to improve approximation of the state distribution at timestep $t$ such that, \eg, the first point of the point set is always assigned to the state with lowest state value.
For more detailed treatments of \arqmc, we refer the reader to \citet{LEcuyer2016-hj} and \citet{Puchhammer2021-og}.

We compare \arqmc (ARQMC) against MC and RQMC on the LQR in \cref{fig:exp-arqmc-full}.
Throughout, we assign RQMC points to states according the state's $\ell_1$-norm, as it is a reasonable proxy for the state's true value and outperformed the $\ell_2$ and $\ell_\infty$ norms in practice.
For policy evaluation (left panel), ARQMC improves upon RQMC, reaching approximately 2.5x lower value estimation error with $4,096$ trajectories.
For policy learning (right panel), ARQMC closely closely matches RQMC but does not provide additional benefit.

We hope those experiments can help motivate research on \arqmc in the context of reinforcement learning.